\documentclass[letterpaper]{article} 
\usepackage{aaai2026}  
\usepackage{times}  
\usepackage{helvet}  
\usepackage{courier}  
\usepackage[hyphens]{url}  
\usepackage{graphicx} 
\usepackage{natbib}  
\usepackage{caption} 
\usepackage{amssymb}
\usepackage{amsmath}
\usepackage{subcaption}
\usepackage{booktabs}
\usepackage{multirow}
\usepackage{xcolor}
\usepackage{enumitem}
\definecolor{teal}{rgb}{0,0.5,0.5}
\usepackage{pythonhighlight}
\usepackage{placeins}
\usepackage{xcolor}

\usepackage{algorithm}
\usepackage{algorithmic}
\usepackage{amsthm}

\theoremstyle{plain}
\newtheorem{definition}{Definition}

\theoremstyle{plain}
\newtheorem{statement}{Statement}

\theoremstyle{remark}
\newtheorem{remark}{Remark}

\theoremstyle{challenge}

\theoremstyle{plain}

\theoremstyle{plain}

\theoremstyle{assumption}

\usepackage{newfloat}
\usepackage{listings}
\DeclareCaptionStyle{ruled}{labelfont=normalfont,labelsep=colon,strut=off} 
\floatstyle{ruled}
\newfloat{listing}{tb}{lst}{}
\floatname{listing}{Listing}
\title{Compensating Distribution Drifts in Class-incremental Learning \\ of Pre-trained Vision Transformers}

\author{
    Xuan Rao \textsuperscript{\rm 1}, Simian Xu \textsuperscript{\rm 2}, Zheng Li \textsuperscript{\rm 3}, Bo Zhao \textsuperscript{\rm 1} \thanks{Corresponding author} \\ Derong Liu \textsuperscript{\rm 4}, Mingming Ha \textsuperscript{\rm 5},
    Cesare Alippi \textsuperscript{\rm 6, 7}
}
\affiliations{
    \textsuperscript{\rm 1} School of Systems Science, Beijing Normal University 
    \ \ \ \textsuperscript{\rm 2} School of Physics, Peking University \\
    \textsuperscript{\rm 3} School of Automation and Intelligent Manufacturing, Southern University of Science and Technology  \\ \textsuperscript{\rm 4} School of Artificial Intelligence, Anhui University \ \
    \textsuperscript{\rm 5} Kuaishou Technology 
    \\ \textsuperscript{\rm 6} Politecnico di Milano \ \ 
    \textsuperscript{\rm 7} Universita' Della Svizzera italiana \\
    \{raoxuan98@mail.bnu.edu.cn, zhaobo@bnu.edu.cn, hamingming@kuaishou.com\}
}

\usepackage{bibentry}

\begin{document}

\maketitle

\begin{abstract}
Recent advances have shown that sequential fine-tuning (SeqFT) of pre-trained vision transformers (ViTs), followed by classifier refinement using approximate distributions of class features, can be an effective strategy for class-incremental learning (CIL). However, this approach is susceptible to distribution drift, caused by the sequential optimization of shared backbone parameters. This results in a mismatch between the distributions of the previously learned classes and that of the updated model, ultimately degrading the effectiveness of classifier performance over time. To address this issue, we introduce a latent space transition operator and propose Sequential Learning with Drift Compensation (SLDC). SLDC aims to align feature distributions across tasks to mitigate the impact of drift. First, we present a linear variant of SLDC, which learns a linear operator by solving a regularized least-squares problem that maps features before and after fine-tuning. Next, we extend this with a weakly nonlinear SLDC variant, which assumes that the ideal transition operator lies between purely linear and fully nonlinear transformations. This is implemented using learnable, weakly nonlinear mappings that balance flexibility and generalization. To further reduce representation drift, we apply knowledge distillation (KD) in both algorithmic variants. Extensive experiments on standard CIL benchmarks demonstrate that SLDC significantly improves the performance of SeqFT. Notably, by combining KD to address representation drift with SLDC to compensate distribution drift, SeqFT achieves performance comparable to joint training across all evaluated datasets. Code: \url{https://github.com/raoxuan98-hash/sldc.git}.


\end{abstract}
%
\section{Introduction}
\begin{figure*}[t]
    \centering
    \includegraphics[width=0.70\linewidth]{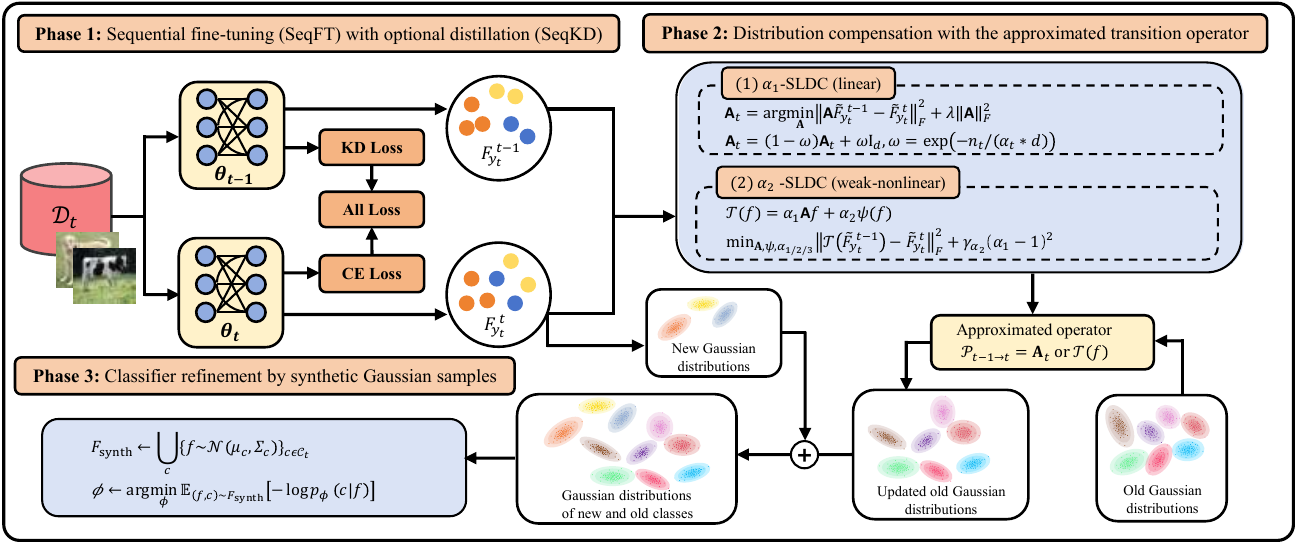}
    \caption{Overview of the SLDC framework. The framework consists of three phases: (1) Sequential fine-tuning with optional distillation (SeqFT/SeqKD); (2) Distribution compensation using an approximated transition operator, either linear ($\alpha_{1}$-SLDC) or weak-nonlinear ($\alpha_2$-SLDC), to align (compensate) previous feature distributions with the new one; (3) Classifier refinement using synthetic Gaussian features sampled from the compensated Gaussian distributions.}
    \label{fig:SLDC_explain}
\end{figure*}

There is a growing interest in applying continual learning (CL) to pre-trained models (PTMs) \cite{dosovitskiy2021an, radford2021learning} by leveraging their rich representations \cite{Zheng_2023_ICCV, Li_2024_WACV, 10882940}. Researchers have shown that sequentially fine-tuning (SeqFT) the backbones of pre-trained vision transformers (ViTs) on downstream tasks, followed by the refinement of the classifier using the approximate distributions of class-wise deep features, offers an effective strategy to class incremental learning (CIL) \cite{Zhang_2023_ICCV, zhang2024slcaunleashpowersequential, 10.1007/978-3-031-73209-6_18}. Notably, unlike methods that introduce task-specific lightweight adaptation to mitigate interference from new tasks \cite{Li_2024_WACV, 10970405}, SeqFT is more computationally efficient, as it eliminates the need of task identification \cite{zhang2024slcaunleashpowersequential, 10.1007/978-3-031-73209-6_18}. 

However, the sequential optimization of shared parameters inevitably introduces representation drifts, which leads to a mismatch between the learned distributions of previous classes and those of the updated model.  

Unlike previous works which mitigate distribution drifts through approaches like distillation, model ensemble, and gradient projection \cite{zhao2024safe, Xiao_2023_CVPR, NEURIPS2024_0f06be00}, our work takes a novel perspective by focusing on compensating for the negative effects of representation drifts once they occur. To this end, we resolve to \textit{model the transformation that occurs in the feature space between consecutive tasks}. In particular, the latent space transition operator that captures how the feature mapping function evolves during task adaptation is defined as:

\begin{definition}[Latent Space Transition Operator]
A latent space transition operator is a mapping $\mathcal{P}_{t-1 \to t}: \mathcal{F}_{t-1} \to \mathcal{F}_t$, where $\mathcal{F}_{t-1}: \mathcal{X} \to \mathbb{R}^d$ and $\mathcal{F}_t: \mathcal{X} \to \mathbb{R}^d$ are (here) neural network-based feature extractors (e.g., backbones of ViTs) that map inputs from the input space $\mathcal{X}$ to a $d$-dimensional feature space at tasks $t-1$ and $t$, respectively.
\end{definition}

Ideally, when the approximate distributions are multivariate Gaussian, the operator \(\mathcal{P}_{t-1 \to t}\) enables the propagation of their first-order (mean) and second-order (covariance) moments from the previous feature space to the new one, which enables consistent classifier refinement despite the representation drift.
However, learning the exact operator $\mathcal{P}_{t-1 \to t}$ would typically require access to the entire input space $\mathcal{X}$ (e.g., the normalized RGB space), which is not available in exemplar-free CIL settings where previous data cannot be preserved. 
To overcome this limitation, we introduce a practical approximation strategy that estimates $\mathcal{P}_{t-1 \to t}$ using only the current task data $\mathcal{D}_t$ and the frozen models $\mathcal{F}_{t-1}$ and $\mathcal{F}_t$.

Accordingly, the Sequential Learning with Drift Compensation (SLDC) method is proposed. First, we propose the $\alpha_{1}$-SLDC method, which learn a linear operator by solving a regularized least-squares problem between the deep features of models $\mathcal{F}_{t-1}$ and $\mathcal{F}_t$ on $\mathcal{D}_t$\footnote{Here, $\alpha/\beta$ denote without or with distillation; subscripts 1/2 denote linear or weak-nonlinear SLDC methods.}. The empirical results show that the linear operator can compensate for the distribution drift appropriately, but it still yields large prediction residuals when predicting the post-optimization deep features, implying that a nonlinear mapping is required. However, the direct implementation of popular nonlinear transformation such as multi‐layer perceptrons (MLPs) leads to overfitting and produces distributions that are less accurate than those obtained with linear operators.


Motivated by these empirical observations, \textbf{\emph{we assume that an ideal operator approximation lies between purely linear and fully nonlinear transformations}}. Correspondingly, we propose the \(\alpha_2\)-SLDC method by constructing a weak-nonlinear transformation to learn the transition operator. 
Building upon \(\alpha_{1,2}\)-SLDC methods, the distillation-enhanced SLDC variants, $\beta_{1,2}$-SLDC, are further developed by constraining model's representation updatings with knowledge distillation (KD).

Notably, the evaluation results show that the combination of distillation (to preserve previous knowledge) with SLDC (to compensate for distribution drifts) enables PTM-based CIL to nearly match the performance of joint training (i.e., training the model using all training data simultaneously), which can be regarded as an empirical upper bound of optimal performance of CIL \cite{sun2025pilot}. It emerges how SLDC achieves near-parity on 10-task CIL scenarios with joint training across two PTMs and four different datasets, with accuracy discrepancies within +0.50\% to -3.29\%, proving the effectiveness of the proposed approach. The novel contributions are:
\begin{enumerate}
\item 

An effective novel CIL methodology is proposedbased on a learned transition operator that models the feature space evolution across successive tasks.

\item Two novel learnable transition operators, the $\alpha_{1}$-SLDC and $\alpha_{2}$-SLDC, along with their distillation-enhanced variants $\beta_{1,2}$-SLDC, are developed based on linear and weak-nonlinear transformations, respectively. The proposed methods can be implemented and integrated with existing approaches.



\end{enumerate}




\section{Methodologies}
\subsection{SeqFT-based CIL with pre-trained ViTs and classifier refinement}
\subsubsection{CIL formalization.} A sequence of training datasets is defined as $\mathcal{D} = \{ \mathcal{D}_1, \ldots, \mathcal{D}_T \}$, where the $t$th dataset is $\mathcal{D}_t = \{(x_{(t,n)}, y_{(t,n)})\}_{n=1}^{n_t}$. Each $\mathcal{D}_t$ contains $n_t$ pairs of input samples $x_{(t,n)} \in \mathcal{X}$ and their corresponding labels $y_{(t,n)} \in \mathcal{Y}_t$, where $\mathcal{X}$ represents the shared input space and $\mathcal{Y}_t$ represents the label space of task $t$. Specifically, $\mathcal{Y}_t \cap \mathcal{Y}_{t'} = \emptyset$ for $t \neq t'$. The cumulative set of observed classes up to task $t$ is denoted as $\mathcal{C}_t = \bigcup_{t'=1}^t \mathcal{Y}_{t'}$.

\subsubsection{ViT architectures.}
The ViT is defined as $\mathcal{G}_{\varphi}\left( x \right) =  \mathcal{C}_\phi \left( \mathcal{F}_{\theta}(x) \right)$, where $\mathcal{F}_{\theta}: \mathcal{X} \to \mathbb{R}^d$ is the pre-trained backbone, $\mathcal{C}_\phi: \mathbb{R}^d \to \mathbb{F}^{|\mathcal{C}_t|}$ is a linear classifier, and $\varphi = \{ \phi, \theta \}$ denotes all trainable parameters \cite{dosovitskiy2021an}. In this paper, we adopt the configuration of SLCA++ and fine-tune the backbones of ViTs using low-rank adaptation (LoRA) \cite{hu2022lora, zhang2024slcaunleashpowersequential}, thus $\theta$ denotes the parameters of the LoRA adapters.

For any label subspace $\mathcal{S} \subseteq \mathcal{C}_t$ (e.g., $\mathcal{Y}_t$ or $\mathcal{C}_t$), the ViT's softmax output is given by
\begin{align}
p_\varphi(x; \mathcal{S})_i = \frac{\exp \left( \left[ \mathcal{G}_\varphi(x) \right]_i \right)}{\sum_{j=1}^{|\mathcal{S}|} \exp \left( \left[ \mathcal{G}_\varphi(x) \right]_j \right)},
\end{align}
where $ i \in \{1, \dots, |\mathcal{S}|\}$. At the task $t$, the model is trained by minimizing the task-specific cross-entropy loss
\begin{align}
   \mathcal{L}_{\text{CE}} \left( \varphi; \mathcal{D}_t \right) = -\frac{1}{B} \sum_{n=1}^{B} \log p_\varphi \left( x_n; \mathcal{Y}_t \right)_{y_n}
\end{align}
where $\left(x_n, y_n \right) \sim \mathcal{D}_{t}$, and $B$ denotes the batch size. 

\subsubsection{Post-hoc classifier refinement.} After the training procedure on task $t$, for each new class $c \in \mathcal{Y}_t$, we assume that its deep features under the PTM mapping $\mathcal{F}_{\theta}$ follow a Gaussian distribution, and its deep feature distribution is approximated by
\begin{align}
	\mu_c &= \frac{1}{n_c} \sum_{i=1}^{n_c} f_c^{(i)}, \label{eq:gaussian_cal_mean} \\
	\Sigma_c &= \frac{1}{n_c} \sum_{i=1}^{n_c} (f_c^{(i)} - \mu_c)(f_c^{(i)} - \mu_c)^\top, \label{eq:gaussian_cal_cov}
\end{align}
where $f_c^{(i)} = \mathcal{F}_\theta(x^{(i)})$ is the feature of sample $x^{(i)}$ with label $c$. Let $\mathcal{H}_{t} =  \left\{ \mathcal{N}(\mu_c, \Sigma_c) \mid c \in \mathcal{C}_t \right\}$ be the set of  all Gaussian distributions up to task $t$. 
The classifier is refined in a post-hoc manner after learning each new task by the synthetic samples from $\mathcal{H}_t$ to improve cross-task decision boundaries as
\begin{align}
\min_{\phi} \mathcal{L}_{\text{CR}} \left( \phi; \mathcal{H}_{t} \right) = -\frac{1}{|\mathcal{C}_t|} \sum_{c \in \mathcal{C}_t} \mathbb{E} \left[ \log p_\phi(f_{\rm synth}; \mathcal{C}_t)_c \right],
\end{align}
where ${f_{\rm synth} \sim \mathcal{N}(\mu_c,\Sigma_c)}$, and $p_\phi(f; \mathcal{C}_t)$ denotes the classifier's softmax output over $\mathcal{C}_{t}$.

\subsection{SLDC for Distribution Drift Compensation }

Figure~\ref{fig:SLDC_explain} provides a visual illustration of SLDC's underlying mechanisms.
Let $\mathcal{F}_{\theta_{t-1}}$ and $\mathcal{F}_{\theta_t}$ be the ViT backbones after training on tasks $t-1$ and $t$, respectively. Given the current task dataset $\mathcal{D}_t$, we define $F_{\mathcal{Y}_t}^{t-1} = \left[ \mathcal{F}_{\theta_{t-1}}(x_{t,1}), \dots, \mathcal{F}_{\theta_{t-1}}(x_{t,n_t}) \right] \in \mathbb{R}^{d \times n_t}$ and $F_{\mathcal{Y}_t}^{t} = \left[ \mathcal{F}_{\theta_{t}}(x_{t,1}), \dots, \mathcal{F}_{\theta_{t}}(x_{t,n_t}) \right] \in \mathbb{R}^{d \times n_t}$ by the feature matrices extracted by backbones $\mathcal{F}_{\theta_{t-1}}$ and $\mathcal{F}_{\theta_{t}}$ on $\mathcal{D}_t$, respectively. 

\subsubsection{Derivation of linear $\alpha_1$-SLDC.} The $\alpha_1$-SLDC estimates a linear transition operator by solving a least-square problem between normalized features. Specifically, let $\tilde{F}_{\mathcal{Y}{t}}^{t-1} \in \mathbb{R}^{d \times n_t}$ and $\tilde{F}_{\mathcal{Y}{t}}^{t} \in \mathbb{R}^{d \times n_t}$ be the column-wise $L_{2}$-normalized versions of $F_{\mathcal{Y}{t}}^{t-1}$ and $F_{\mathcal{Y}_{t}}^{t}$, respectively. The linear operator $\mathbf{A}_{t} \in \mathbb{R}^{d \times d}$ for approximating $\mathcal{P}_{t-1 \rightarrow t}$ is obtained by solving the regularized least-square solution 
\begin{align}
	\mathbf{A}_{t} & = \arg\min_{\mathbf{A}} \| \mathbf{A} \tilde{F}_{\mathcal{Y}_t}^{t-1} - \tilde{F}_{\mathcal{Y}_t}^{t} \|_F^2 + \gamma_{\alpha_{1}} \| \mathbf{A} \|_{F}^2 \\
	& = \tilde{F}_{\mathcal{Y}_t}^{t}\big(\tilde{F}_{\mathcal{Y}_t}^{t-1}\big)^\top
	\Bigl(\tilde{F}_{\mathcal{Y}_t}^{t-1}\big( \tilde{F}_{\mathcal{Y}_t}^{t-1}\big)^\top + \gamma_{\alpha_{1}} I_d\Bigr)^{-1},
\end{align}
where $\gamma_{\alpha_{1}}$ is the regularization coefficient and $I_{d} \in \mathbb{R}^{d\times d}$ is the identity matrix. 
In addition, there are some cases where the number of task-specific samples \( n_t \) is too small to obtain a robust estimation of the linear operator.
To avoid this problem, we regularize \( \mathbf{A}_{t} \) by a heuristic re-weighting process based on sample complexity as
\begin{align}
\mathbf{A}_{t} = (1 - w) \mathbf{A}_{t} + w I_d,
\end{align}
where $ w = \exp \left( -\frac{ n_{t}} {\alpha_{\rm temp} d} \right)$ and $\alpha_{\rm temp}$ are the weighting and temperature coefficients, respectively. 

Once $\mathbf{A}_{t}$ is obtained, the Gaussian distributions of previous tasks' classes $c \in \mathcal{C}_{t-1}$ are compensated by
\begin{align}
\mu_{c} \leftarrow \mathbf{A}_{t} \mu_{c} \quad \Sigma_{c}\leftarrow  \mathbf{A}_{t} \Sigma_{c}\mathbf{A}_{t}^\top.
\end{align}
This process is applied recursively as new tasks arrive. In Statement \ref{theorm:LS} of the appendix, it is proved that the above updating formulations follow the close-formed solution to the linear transformation of a Gaussian distribution. 

\subsubsection{Derivation of weak-nonlinear $\alpha_{2}$-SLDC.} 
Although the task-wise linear operator $\mathbf{A}_{t}$ in $\alpha_{1}$-SLDC can mitigate distribution drifts to some extent, residual errors between the predicted and actual features still remain. While nonlinear MLPs could address the under-fitting problem, they suffer from over-fitting and yield less accurate transformed distributions than linear transformations. 

Based on these empirical observations, we assume that the ideal transition operator $P_{t-1 \rightarrow t}$ for SeqFT-based CIL with pre-trained ViTs resides between purely linear and fully nonlinear transformations, i.e., $P_{t-1 \rightarrow t}$ is weak-nonlinear.


Motivated by the assumption, the $\alpha_{2}$-SLDC is proposed by defining the weak-nonlinear transformation
\begin{align}
   \mathcal{T}\left( f \right) =  c_{1} \mathbf{A} f + c_{2} \psi \left( f \right).
\end{align}
Specifically, $c_{1/2}$ are learnable contribution coefficients which satisfies $c_{1/2} \ge 0 $ and $c_{1} + c_{2} = 1$. In particular, we instantiate $\mathbf{A}$ as a learnable matrix and $\psi(f)$ as a two-layer MLP with ReLU activation. To optimize $\mathcal{T}(f)$, a regularized optimization objective is defined by
\begin{align}
   \min_{\mathbf{A}, \psi, c_{1/2}} & \left\| \mathcal{T} \left( \tilde{F}_{\mathcal{Y}_{t}}^{t-1} \right) - \tilde{F}_{\mathcal{Y}_{t}}^{t} \right\|_F^2  + \gamma_{\alpha_{2}} (c_1 - 1)^2,
\end{align}
where $\gamma_{\alpha_{2}} (c_{1} - 1)^2$ is the regularization term controlling the contribution of nonlinear $\psi \left(f\right)$. 

In practice, the optimization process for $\mathcal{T}(f)$ is end-to-end by the gradient optimizer, and the training details are presented in the experiment section. Specifically, in Statements \ref{state:affinity} and \ref{theorem:training}, some theoretical claims on the characteristics of transition operator are given based on the neural tangent kernel (NTK) theory \citep{ntk_th}. 

After obtaining the weak-nonlinear transformation $\mathcal{T}(f)$, the Monte Carlo sampling is used to estimate the updated Gaussian distributions for previous classes $c \in \mathcal{C}_{t-1}$. Specifically, for each class $c$, we generate $N \gg d$ synthetic samples from its original Gaussian distribution $\mathcal{N}(\mu_c, \Sigma_c)$
\begin{align}
	f_c^{(i)} \sim \mathcal{N}(\mu_c, \Sigma_c), \quad i = 1,\dots,N
\end{align}
These samples are then compensated by the weak-nonlinear transformation as
\begin{align}
    \tilde{f}_c^{(i)} = \mathcal{T}(f_c^{(i)}), \quad i = 1,\dots,N
    \label{eq:alpha_2-SLDC}
\end{align}
Hereafter, the mean $\mu_{c}$ and covariance $\Sigma_{c}$ for $c \in \mathcal{C}_{t-1}$ are compensated by re-calculating \eqref{eq:gaussian_cal_mean} and \eqref{eq:gaussian_cal_cov} using the transformed samples in \eqref{eq:alpha_2-SLDC}. Finally, distributions of old classes in $\mathcal{H}_t$ are replaced by the updated ones before executing classifier refinement.

\subsubsection{Distillation-enhanced SLDC variants.} Typically, the unconstrained optimization for ViT backbones makes the performance of SeqFT for CIL sensitive to several hyper-parameters such as batch size, learning rate and tuning epochs. Considering these issues, the distillation-enhanced variants of $\alpha_{1,2}$-SLDC are proposed by incorporating a feature-based distillation loss, i.e., 
\begin{align}
	\mathcal{L}_{\rm KD} = -\frac{1}{B} \sum_{n=1}^{B} \Vert \mathcal{F}_{\theta_{t-1}}\left( x_{n} \right) - \mathcal{F}_{_{\theta}} \left( x_{n} \right) \Vert^2,
\end{align}
In addition, a regularization loss is considered to maintain the $L_2$-norm of feature vectors as
\begin{align}
	\mathcal{L}_{\rm Norm} = -\frac{1}{B} \sum_{n=1}^{B} \left( \Vert \mathcal{F}_{\theta_{t-1}}\left( x_{n} \right) \Vert - \Vert  \mathcal{F}_{_{\theta}} \left( x_{n} \right) \Vert \right)^2,
\end{align}
Consequently, the overall loss for optimizing the ViT backbone in $\beta$-SLDC is 
\begin{align}
	 \mathcal{L}_{\rm All} = \mathcal{L}_{\rm CE}  + \gamma_{\rm kd} \mathcal{L}_{\rm KD} + \gamma_{\rm norm} \mathcal{L}_{\rm Norm},
\end{align}
where $\gamma_{\rm KD}$ and $\gamma_{\rm Norm}$ are the balance coefficients. In particular, we refer $\beta_{1,2}$-SLDC to the distillation-enhanced $\alpha_{1,2}$-SLDC variants, respectively. We also refer SeqKD to the distillation-enhanced SeqFT in the following sections. 
\subsubsection{Improved operator estimation with auxiliary unlabeled data.} 
In certain scenarios, limited dataset size and insufficient sample diversity can lead to inaccurate approximations of the transfer operator. To address this challenge, this paper proposes auxiliary data enrichment (ADE) to improve the prediction by leveraging unlabeled auxiliary data from arbitrary sources. \emph{Crucially, ADE operates without requiring labeled data and remains consistent with the exemplar-free continual learning (CIL) framework since it does not preserve any task-relevant data from previous tasks.}

\section{Related Works}
\begin{table*}[t]
\centering
\caption{State-of-the-art CIL performance comparison across CUB-200, Cars-196, CIFAR-100, and ImageNet-R by a self-supervised pre-trained ViT-B/16 with the MoCo-V3 approach.}
\scalebox{0.55}{
    \begin{tabular}{lccccccccc}
        \toprule
        \multirow{2}{*}{Method} & & \multicolumn{2}{c}{CUB-200} & \multicolumn{2}{c}{Cars-196} & \multicolumn{2}{c}{CIFAR-100} & \multicolumn{2}{c}{ImageNet-R} \\ 
        \cmidrule(lr){3-4} \cmidrule(lr){5-6} \cmidrule(lr){7-8} \cmidrule(lr){9-10}
        & & Last-Acc & Inc-Acc & Last-Acc & Inc-Acc & Last-Acc & Inc-Acc & Last-Acc & Inc-Acc \\
        \midrule
        Joint-Training & {} & 81.82{\scriptsize$\pm$0.29} & - & 81.16{\scriptsize$\pm$0.06} & - & 88.86{\scriptsize$\pm$0.14} & - & 75.95{\scriptsize$\pm$0.23} & - \\
        \midrule
        BiC & {} & 74.39{\scriptsize$\pm$1.12} & 82.13{\scriptsize$\pm$0.33} & 65.57{\scriptsize$\pm$0.93} & 73.95{\scriptsize$\pm$0.29} & 80.57{\scriptsize$\pm$0.86} & 89.39{\scriptsize$\pm$0.33} & 57.36{\scriptsize$\pm$2.68} & 68.07{\scriptsize$\pm$0.22} \\
        LwF & {} & 61.66{\scriptsize$\pm$1.95} & 73.90{\scriptsize$\pm$1.91} & 52.45{\scriptsize$\pm$0.48} & 63.87{\scriptsize$\pm$0.31} & 77.94{\scriptsize$\pm$1.00} & 86.90{\scriptsize$\pm$0.90} & 60.74{\scriptsize$\pm$0.30} & 68.55{\scriptsize$\pm$0.65} \\
        RanPAC & {} & 74.43{\scriptsize$\pm$0.43} & 83.63{\scriptsize$\pm$0.01} & 63.21{\scriptsize$\pm$0.02} & 74.01{\scriptsize$\pm$0.47} & 86.47{\scriptsize$\pm$0.52} & 90.81{\scriptsize$\pm$1.05} & 69.11{\scriptsize$\pm$0.69} & 75.20{\scriptsize$\pm$0.34} \\
        SLCA & {} & 73.01{\scriptsize$\pm$0.16} & 82.13{\scriptsize$\pm$0.34} & 66.04{\scriptsize$\pm$0.08} & 72.59{\scriptsize$\pm$0.04} & 85.27{\scriptsize$\pm$0.08} & 89.51{\scriptsize$\pm$1.04} & 68.07{\scriptsize$\pm$0.21} & 73.04{\scriptsize$\pm$0.56} \\
        SLCA++ & {} & 75.48{\scriptsize$\pm$0.31} & 82.94{\scriptsize$\pm$0.73} & 69.71{\scriptsize$\pm$0.10} & 75.67{\scriptsize$\pm$0.32} & 84.77{\scriptsize$\pm$0.18} & 89.53{\scriptsize$\pm$0.98} & 69.01{\scriptsize$\pm$0.42} & 74.75{\scriptsize$\pm$0.69} \\
        CoMA  & {} & 75.12{\scriptsize$\pm$0.27} & 82.76{\scriptsize$\pm$0.16} & 67.48{\scriptsize$\pm$0.19} & 74.90{\scriptsize$\pm$0.87} & 86.59{\scriptsize$\pm$0.51} & 91.02{\scriptsize$\pm$0.47} & 69.33{\scriptsize$\pm$0.22} & 75.64{\scriptsize$\pm$0.13} \\
        CoFiMA & {} & 77.65{\scriptsize$\pm$0.18} & 83.54{\scriptsize$\pm$0.16} & 69.51{\scriptsize$\pm$0.16} & 76.21{\scriptsize$\pm$0.83} & 87.44{\scriptsize$\pm$0.47} & 91.13{\scriptsize$\pm$0.53} & 70.87{\scriptsize$\pm$0.31} & 76.09{\scriptsize$\pm$0.78} \\
            
        \midrule
        SeqFT & {} & 64.40{\scriptsize$\pm$1.65} & 77.77{\scriptsize$\pm$0.61} & 60.42{\scriptsize$\pm$1.50} & 72.12{\scriptsize$\pm$0.63} & 73.36{\scriptsize$\pm$0.90} & 80.40{\scriptsize$\pm$2.01} & 61.37{\scriptsize$\pm$0.25} & 70.55{\scriptsize$\pm$0.55} \\
        SeqFT + MLPDC & {} & 70.56{\scriptsize$\pm$1.09} $^{\textcolor{red}{\uparrow 6.16}}$ & 82.70{\scriptsize$\pm$0.72} & 67.87{\scriptsize$\pm$0.51} $^{\textcolor{red}{\uparrow 7.45}}$ & 79.68{\scriptsize$\pm$0.57} & 79.21{\scriptsize$\pm$1.44} $^{\textcolor{red}{\uparrow 5.85}}$ & 86.98{\scriptsize$\pm$0.86} & 69.88{\scriptsize$\pm$0.31} $^{\textcolor{red}{\uparrow 8.51}}$ & 76.71{\scriptsize$\pm$0.56} \\
        $\alpha_{1}$-SLDC (ours) & {} & 70.42{\scriptsize$\pm$1.01} $^{\textcolor{red}{\uparrow 6.02}}$ & 82.86{\scriptsize$\pm$0.85} & 61.01{\scriptsize$\pm$0.74} $^{\textcolor{red}{\uparrow 0.59}}$ & 76.33{\scriptsize$\pm$0.57} & 79.84{\scriptsize$\pm$1.12} $^{\textcolor{red}{\uparrow 6.48}}$ & 88.15{\scriptsize$\pm$0.75} & 71.81{\scriptsize$\pm$0.39} $^{\textcolor{red}{\uparrow 10.44}}$ & 77.73{\scriptsize$\pm$0.43} \\
        $\alpha_{2}$-SLDC (ours) & {} & 78.98{\scriptsize$\pm$0.95} $^{\textcolor{red}{\uparrow 14.58}}$ & 86.70{\scriptsize$\pm$0.72} & 77.53{\scriptsize$\pm$0.05} $^{\textcolor{red}{\uparrow 17.11}}$ & 84.25{\scriptsize$\pm$0.52} & 81.75{\scriptsize$\pm$0.74} $^{\textcolor{red}{\uparrow 8.39}}$ & 88.75{\scriptsize$\pm$0.79} & 71.38{\scriptsize$\pm$0.40} $^{\textcolor{red}{\uparrow 10.01}}$ & 77.79{\scriptsize$\pm$0.39} \\
        SeqFT + MLPDC + ADE  & {} & 76.66{\scriptsize$\pm$1.22} $^{\textcolor{red}{\uparrow 12.26}}$ & 85.74{\scriptsize$\pm$0.91} & 74.24{\scriptsize$\pm$0.47} $^{\textcolor{red}{\uparrow 13.82}}$ & 82.90{\scriptsize$\pm$0.42} & 79.65{\scriptsize$\pm$0.93} $^{\textcolor{red}{\uparrow 6.29}}$ & 86.94{\scriptsize$\pm$0.99} & 70.54{\scriptsize$\pm$0.72} $^{\textcolor{red}{\uparrow 9.17}}$ & 77.04{\scriptsize$\pm$0.40} \\
        $\alpha_{1}$-SLDC + ADE (ours)  & {} & 78.03{\scriptsize$\pm$1.36} $^{\textcolor{red}{\uparrow 13.63}}$ & 86.54{\scriptsize$\pm$0.80} & 76.26{\scriptsize$\pm$0.59} $^{\textcolor{red}{\uparrow 15.84}}$ & 83.87{\scriptsize$\pm$0.40} & 81.57{\scriptsize$\pm$0.98} $^{\textcolor{red}{\uparrow 8.21}}$ & 88.78{\scriptsize$\pm$0.75} & 72.29{\scriptsize$\pm$0.42} $^{\textcolor{red}{\uparrow 10.92}}$ & 77.95{\scriptsize$\pm$0.31} \\
        $\alpha_{2}$-SLDC  + ADE (ours) & {} & 79.43{\scriptsize$\pm$0.77} $^{\textcolor{red}{\uparrow 15.03}}$ & 86.92{\scriptsize$\pm$0.88} & 77.51{\scriptsize$\pm$0.21} $^{\textcolor{red}{\uparrow 17.09}}$ & 84.32{\scriptsize$\pm$0.44} & 83.15{\scriptsize$\pm$0.81} $^{\textcolor{red}{\uparrow 9.79}}$ & 89.26{\scriptsize$\pm$0.82} & 72.47{\scriptsize$\pm$0.08} $^{\textcolor{red}{\uparrow 11.10}}$ & 77.95{\scriptsize$\pm$0.28} \\
        \midrule
        SeqKD & {} & 76.97{\scriptsize$\pm$0.20} & 86.00{\scriptsize$\pm$0.66} & 73.87{\scriptsize$\pm$0.66} & 82.37{\scriptsize$\pm$0.68} & 80.35{\scriptsize$\pm$0.41} & 88.09{\scriptsize$\pm$0.92} & 66.93{\scriptsize$\pm$0.28} & 75.07{\scriptsize$\pm$0.45} \\
        SeqKD + MLPDC & {} & 72.56{\scriptsize$\pm$0.81} $^{\textcolor{red}{\downarrow 4.41}}$ & 83.44{\scriptsize$\pm$0.86} & 71.18{\scriptsize$\pm$0.37} $^{\textcolor{red}{\downarrow 2.69}}$ & 81.07{\scriptsize$\pm$0.51} & 82.59{\scriptsize$\pm$0.95} $^{\textcolor{red}{\uparrow 2.24}}$ & 88.80{\scriptsize$\pm$1.01} & 72.11{\scriptsize$\pm$0.22} $^{\textcolor{red}{\uparrow 5.18}}$ & 77.44{\scriptsize$\pm$0.41} \\
        $\beta_{1}$-SLDC (ours) & {} & 80.55{\scriptsize$\pm$0.53} $^{\textcolor{red}{\uparrow 3.58}}$ & 87.29{\scriptsize$\pm$0.76} & 77.79{\scriptsize$\pm$0.27} $^{\textcolor{red}{\uparrow 3.92}}$ & 84.19{\scriptsize$\pm$0.43} & 85.50{\scriptsize$\pm$0.53} $^{\textcolor{red}{\uparrow 5.15}}$ & 90.52{\scriptsize$\pm$0.97} & 73.00{\scriptsize$\pm$0.13} $^{\textcolor{red}{\uparrow 6.07}}$ & 78.08{\scriptsize$\pm$0.25} \\
        $\beta_{2}$-SLDC (ours) & {} & 81.82{\scriptsize$\pm$0.52} $^{\textcolor{red}{\uparrow 4.85}}$ & 87.60{\scriptsize$\pm$0.71} & 80.10{\scriptsize$\pm$0.31} $^{\textcolor{red}{\uparrow 6.23}}$ & 85.07{\scriptsize$\pm$0.54} & 85.16{\scriptsize$\pm$0.29} $^{\textcolor{red}{\uparrow 4.81}}$ & 90.30{\scriptsize$\pm$0.96} & 73.01{\scriptsize$\pm$0.11} $^{\textcolor{red}{\uparrow 6.08}}$ & 77.96{\scriptsize$\pm$0.28} \\
        SeqKD + MLPDC + ADE  & {} & 80.54{\scriptsize$\pm$0.49} $^{\textcolor{red}{\uparrow 3.57}}$ & 87.26{\scriptsize$\pm$0.80} & 78.77{\scriptsize$\pm$0.28} $^{\textcolor{red}{\uparrow 4.90}}$ & 84.53{\scriptsize$\pm$0.42} & 82.42{\scriptsize$\pm$0.81} $^{\textcolor{red}{\uparrow 2.07}}$ & 88.70{\scriptsize$\pm$0.97} & 71.11{\scriptsize$\pm$0.25} $^{\textcolor{red}{\uparrow 4.18}}$ & 77.06{\scriptsize$\pm$0.29} \\
        $\beta_{1}$-SLDC + ADE (ours)  & {} & 82.21{\scriptsize$\pm$0.53} $^{\textcolor{red}{\uparrow 5.24}}$ & 87.85{\scriptsize$\pm$0.68} & 80.59{\scriptsize$\pm$0.29} $^{\textcolor{red}{\uparrow 6.72}}$ & 85.31{\scriptsize$\pm$0.37} & 86.02{\scriptsize$\pm$0.31} $^{\textcolor{red}{\uparrow 5.67}}$ & 90.62{\scriptsize$\pm$0.94} & 73.42{\scriptsize$\pm$0.11} $^{\textcolor{red}{\uparrow 6.49}}$ & 78.05{\scriptsize$\pm$0.33} \\
        $\beta_{2}$-SLDC  + ADE (ours) & {} & 82.32{\scriptsize$\pm$0.57} $^{\textcolor{red}{\uparrow 5.35}}$ & 87.78{\scriptsize$\pm$0.76} & 80.61{\scriptsize$\pm$0.31} $^{\textcolor{red}{\uparrow 6.74}}$ & 85.32{\scriptsize$\pm$0.42} & 86.12{\scriptsize$\pm$0.23} $^{\textcolor{red}{\uparrow 5.77}}$ & 90.52{\scriptsize$\pm$0.98} & 73.14{\scriptsize$\pm$0.22} $^{\textcolor{red}{\uparrow 6.21}}$ & 77.96{\scriptsize$\pm$0.28} \\
        \bottomrule
    \end{tabular}}
\label{tab:results_mocov3}
\end{table*}

Based on strategies for dealing with representation drifts, existing research on ViT-based CIL approaches can be divided into four types. 

The first category optimizes task-specific adapters for each new task and selects appropriate adapters during inference based on the characteristics of test samples \cite{10970405, Li_2024_WACV}. Typically, these methods decompose the prediction process into two hierarchical stages, i.e., the task identity prediction and the within-task label prediction using the corresponding adapter. 
However, these methods rely heavily on task identity prediction accuracy, incur high computational overhead due to repeated forward passes, and face linearly scaling storage demands for adapters.

The second approach trains a shared backbone or lightweight adapter across tasks by using techniques like reduced learning rates, distillation, model merging, or gradient projection to mitigate catastrophic forgetting~\cite{Zhang_2023_ICCV, Gao_2023_ICCV, 10.1007/978-3-031-73209-6_18, NEURIPS2024_0f06be00}. For example, slow learner with classifier alignment (SLCA) adapts ViT backbones with lower learning rates to preserve pre-trained knowledge \cite{Zhang_2023_ICCV}. Enhancements like continual model averaging (CoMA) and continual fisher-weighted model averaging (CoFiMA) improve SLCA by averaging current and past models \cite{10.1007/978-3-031-73209-6_18}, which proportionally average current and past models to enhance SLCA's performance. SLCA++ further integrates lightweight adapters in SLCA, and achieves comparable results with minimal parameter optimization \cite{zhang2024slcaunleashpowersequential}. However, these methods remain vulnerable to representation drifts from progressive optimization.

The third approach combines multiple shared adapters with instance-level feature adaptation. Learning to prompt (L2P) uses a fixed prompt pool and learnable query vectors to dynamically select prompts based on sample features \cite{Wang_2022_CVPR}. DualPrompt extends L2P with supplementary task-specific prompts \cite{wang2022dualprompt}, while CODA-Prompt employs an input-dependent key-value mechanism to achieve finer-grained prompts \cite{Smith_2023_CVPR}.

The fourth category freezes PTMs and leverages the pre-trained features only. First session adaptation (FSA) optimizes PTMs only in the first task and applies exemplar-free CIL by incremental linear discriminant analysis (LDA) \cite{Panos_2023_ICCV}. RanPAC enhances FSA by projecting ViT features into a 10,000-dimensional space with a nonlinear ReLU mapping \cite{NEURIPS2023_2793dc35}. LayUP enhances RanPAC's performance by concatenating outputs from multiple feature layers \cite{ahrens2024read}. 

Beyond PTM-based CIL, there were methods compensating the distribution drifts during CIL \cite{yu2020semantic, gomez2024exemplar}. For example, AddGauss tackles task-recency bias by adapting class covariance matrices with nonlinear mappings \cite{NEURIPS2024_73ba81c7}. Meanwhile, DPCR quantifies feature space semantic drifts using linear task-wise semantic drift projections and categorical information projections \cite{dpcr}, DS-AL constructs an analytic incremental classifier based on the recursive least-squares method \cite{zhuang2024ds}. Notably, SLDC methods take insights from AddGauss and investigate the efficacy of linear, weak-nonlinear, and nonlinear transformations in the context of PTM-based CIL research. 

\section{Experiment Evaluations}
\begin{table*}[htbp]
	\centering
	\caption{State-of-the-art CIL performance comparison across CUB-200, Cars-196, CIFAR-100, and ImageNet-R by a supervisedly pre-trained ViT-B/16 on ImageNet-21K.}
	\scalebox{0.55}{
		\begin{tabular}{lccccccccc}
			\toprule
			\multirow{2}{*}{Method} & & \multicolumn{2}{c}{CUB-200} & \multicolumn{2}{c}{Cars-196} & \multicolumn{2}{c}{CIFAR-100} & \multicolumn{2}{c}{ImageNet-R} \\ 
			\cmidrule(lr){3-4} \cmidrule(lr){5-6} \cmidrule(lr){7-8} \cmidrule(lr){9-10}
			& & Last-Acc & Inc-Acc & Last-Acc & Inc-Acc & Last-Acc & Inc-Acc & Last-Acc & Inc-Acc \\
			\midrule
			Joint-Training & {} & 88.43{\scriptsize$\pm$0.25} & - & 83.79{\scriptsize$\pm$0.25} & - & 93.56{\scriptsize$\pm$0.17} & - & 82.74{\scriptsize$\pm$0.14} & - \\
			\midrule
			BiC & {} & 81.91{\scriptsize$\pm$2.50} & 89.29{\scriptsize$\pm$1.57} & 63.10{\scriptsize$\pm$5.71} & 73.75{\scriptsize$\pm$2.37} & 88.45{\scriptsize$\pm$0.57} & 93.37{\scriptsize$\pm$0.32} & 64.89{\scriptsize$\pm$0.80} & 73.66{\scriptsize$\pm$1.61} \\
			LwF & {} & 69.75{\scriptsize$\pm$1.37} & 80.45{\scriptsize$\pm$2.08} & 49.94{\scriptsize$\pm$3.24} & 63.28{\scriptsize$\pm$1.11} & 87.99{\scriptsize$\pm$0.05} & 92.13{\scriptsize$\pm$1.16} & 67.29{\scriptsize$\pm$1.67} & 74.47{\scriptsize$\pm$1.48} \\
			RanPAC & {} & 85.82{\scriptsize$\pm$0.53} & 91.47{\scriptsize$\pm$0.96} & 53.84{\scriptsize$\pm$0.84} & 64.39{\scriptsize$\pm$1.18} & 90.09{\scriptsize$\pm$0.25} & 93.31{\scriptsize$\pm$0.98} & 72.62{\scriptsize$\pm$0.11} & 78.35{\scriptsize$\pm$0.58} \\
			SLCA & {} & 84.71{\scriptsize$\pm$0.40} & 90.94{\scriptsize$\pm$0.68} & 67.73{\scriptsize$\pm$0.85} & 76.93{\scriptsize$\pm$1.21} & 91.53{\scriptsize$\pm$0.28} & 94.09{\scriptsize$\pm$0.87} & 77.00{\scriptsize$\pm$0.33} & 81.17{\scriptsize$\pm$0.64} \\
            SLCA++ & {} & 86.59{\scriptsize$\pm$0.29} & 91.63{\scriptsize$\pm$0.72} & 73.97{\scriptsize$\pm$0.22} & 79.49{\scriptsize$\pm$0.80} & 91.46{\scriptsize$\pm$0.18} & 94.20{\scriptsize$\pm$0.71} & 78.09{\scriptsize$\pm$0.22} & 82.95{\scriptsize$\pm$0.78} \\
			CoMA  & {} & 85.95{\scriptsize$\pm$0.29} & 90.75{\scriptsize$\pm$0.39} & 73.35{\scriptsize$\pm$0.50} & 78.55{\scriptsize$\pm$0.42} & 92.00{\scriptsize$\pm$0.13} & 94.12{\scriptsize$\pm$0.63} & 77.47{\scriptsize$\pm$0.05} & 81.32{\scriptsize$\pm$0.17} \\
			CoFiMA & {} & 87.11{\scriptsize$\pm$0.56} & 91.87{\scriptsize$\pm$0.69} & 76.96{\scriptsize$\pm$0.64} & 82.65{\scriptsize$\pm$0.96} & 92.77{\scriptsize$\pm$0.24} & 94.89{\scriptsize$\pm$0.94} & 78.25{\scriptsize$\pm$0.26} & 81.48{\scriptsize$\pm$0.56} \\
            \midrule
            SeqFT & {} & 76.57{\scriptsize$\pm$1.62} & 85.84{\scriptsize$\pm$0.47} & 54.53{\scriptsize$\pm$1.75} & 69.48{\scriptsize$\pm$0.83} & 83.14{\scriptsize$\pm$1.37} & 88.06{\scriptsize$\pm$1.03} & 68.56{\scriptsize$\pm$0.94} & 77.46{\scriptsize$\pm$0.31} \\
            SeqFT + MLPDC & {} & 68.32{\scriptsize$\pm$1.79} $^{\textcolor{red}{\downarrow 8.25}}$ & 84.29{\scriptsize$\pm$0.95} & 64.65{\scriptsize$\pm$0.41} $^{\textcolor{red}{\uparrow 10.12}}$ & 78.54{\scriptsize$\pm$0.43} & 87.20{\scriptsize$\pm$1.00} $^{\textcolor{red}{\uparrow 4.06}}$ & 91.96{\scriptsize$\pm$0.57} & 73.38{\scriptsize$\pm$0.30} $^{\textcolor{red}{\uparrow 4.82}}$ & 81.45{\scriptsize$\pm$0.65} \\
            $\alpha_{1}$-SLDC (ours) & {} & 71.49{\scriptsize$\pm$2.54} $^{\textcolor{red}{\downarrow 5.08}}$ & 84.65{\scriptsize$\pm$1.01} & 46.78{\scriptsize$\pm$1.80} $^{\textcolor{red}{\downarrow 7.75}}$ & 68.64{\scriptsize$\pm$1.34} & 87.45{\scriptsize$\pm$1.09} $^{\textcolor{red}{\uparrow 4.31}}$ & 92.41{\scriptsize$\pm$0.50} & 76.85{\scriptsize$\pm$0.20} $^{\textcolor{red}{\uparrow 8.29}}$ & 82.85{\scriptsize$\pm$0.57} \\
            $\alpha_{2}$-SLDC (ours) & {} & 78.65{\scriptsize$\pm$2.18} $^{\textcolor{red}{\uparrow 2.08}}$ & 88.72{\scriptsize$\pm$1.01} & 74.07{\scriptsize$\pm$0.78} $^{\textcolor{red}{\uparrow 19.54}}$ & 83.32{\scriptsize$\pm$0.55} & 88.69{\scriptsize$\pm$0.44} $^{\textcolor{red}{\uparrow 5.55}}$ & 93.02{\scriptsize$\pm$0.56} & 77.05{\scriptsize$\pm$0.04} $^{\textcolor{red}{\uparrow 8.49}}$ & 82.96{\scriptsize$\pm$0.46} \\
            SeqFT + MLPDC + ADE  & {} & 75.44{\scriptsize$\pm$1.71} $^{\textcolor{red}{\downarrow 1.13}}$ & 86.69{\scriptsize$\pm$1.09} & 69.26{\scriptsize$\pm$0.47} $^{\textcolor{red}{\uparrow 14.73}}$ & 81.15{\scriptsize$\pm$0.32} & 88.44{\scriptsize$\pm$0.48} $^{\textcolor{red}{\uparrow 5.30}}$ & 92.21{\scriptsize$\pm$0.82} & 76.98{\scriptsize$\pm$0.15} $^{\textcolor{red}{\uparrow 8.42}}$ & 82.73{\scriptsize$\pm$0.44} \\
            $\alpha_{1}$-SLDC + ADE (ours) & {} & 77.02{\scriptsize$\pm$2.34} $^{\textcolor{red}{\uparrow 0.45}}$ & 87.93{\scriptsize$\pm$0.97} & 73.01{\scriptsize$\pm$0.97} $^{\textcolor{red}{\uparrow 18.48}}$ & 82.82{\scriptsize$\pm$0.52} & 88.73{\scriptsize$\pm$0.86} $^{\textcolor{red}{\uparrow 5.59}}$ & 92.92{\scriptsize$\pm$0.45} & 78.14{\scriptsize$\pm$0.10} $^{\textcolor{red}{\uparrow 9.58}}$ & 83.38{\scriptsize$\pm$0.45} \\
            $\alpha_{2}$-SLDC  + ADE (ours)  & {} & 77.56{\scriptsize$\pm$2.00} $^{\textcolor{red}{\uparrow 0.99}}$ & 88.20{\scriptsize$\pm$0.94} & 73.04{\scriptsize$\pm$0.57} $^{\textcolor{red}{\uparrow 18.51}}$ & 83.02{\scriptsize$\pm$0.44} & 89.83{\scriptsize$\pm$0.53} $^{\textcolor{red}{\uparrow 6.69}}$ & 93.43{\scriptsize$\pm$0.61} & 78.82{\scriptsize$\pm$0.26} $^{\textcolor{red}{\uparrow 10.26}}$ & 83.61{\scriptsize$\pm$0.36} \\
            \midrule
            SeqKD & {} & 86.75{\scriptsize$\pm$0.29} & 92.22{\scriptsize$\pm$0.55} & 75.62{\scriptsize$\pm$0.32} & 83.36{\scriptsize$\pm$0.63} & 88.03{\scriptsize$\pm$0.62} & 92.85{\scriptsize$\pm$0.91} & 74.04{\scriptsize$\pm$0.38} & 81.25{\scriptsize$\pm$0.32} \\
            SeqKD + MLPDC & {} & 75.76{\scriptsize$\pm$1.23} $^{\textcolor{red}{\downarrow 10.99}}$ & 87.22{\scriptsize$\pm$0.60} & 70.19{\scriptsize$\pm$0.65} $^{\textcolor{red}{\downarrow 5.43}}$ & 80.69{\scriptsize$\pm$0.86} & 89.65{\scriptsize$\pm$0.56} $^{\textcolor{red}{\uparrow 1.62}}$ & 93.26{\scriptsize$\pm$0.75} & 78.57{\scriptsize$\pm$0.17} $^{\textcolor{red}{\uparrow 4.53}}$ & 83.27{\scriptsize$\pm$0.73} \\
            $\beta_{1}$-SLDC (ours) & {} & 83.76{\scriptsize$\pm$1.41} $^{\textcolor{red}{\downarrow 2.99}}$ & 91.06{\scriptsize$\pm$0.72} & 73.71{\scriptsize$\pm$1.03} $^{\textcolor{red}{\downarrow 1.91}}$ & 82.48{\scriptsize$\pm$0.71} & 91.21{\scriptsize$\pm$0.45} $^{\textcolor{red}{\uparrow 3.18}}$ & 94.27{\scriptsize$\pm$0.69} & 79.56{\scriptsize$\pm$0.44} $^{\textcolor{red}{\uparrow 5.52}}$ & 83.82{\scriptsize$\pm$0.55} \\
            $\beta_{2}$-SLDC (ours) & {} & 85.85{\scriptsize$\pm$0.49} $^{\textcolor{red}{\downarrow 0.90}}$ & 91.92{\scriptsize$\pm$0.60} & 79.91{\scriptsize$\pm$0.47} $^{\textcolor{red}{\uparrow 4.29}}$ & 85.11{\scriptsize$\pm$0.49} & 90.98{\scriptsize$\pm$0.27} $^{\textcolor{red}{\uparrow 2.95}}$ & 94.20{\scriptsize$\pm$0.72} & 79.54{\scriptsize$\pm$0.02} $^{\textcolor{red}{\uparrow 5.50}}$ & 83.96{\scriptsize$\pm$0.48} \\
            SeqKD + MLPDC + ADE & {} & 85.05{\scriptsize$\pm$0.80} $^{\textcolor{red}{\downarrow 1.70}}$ & 91.43{\scriptsize$\pm$0.76} & 77.30{\scriptsize$\pm$0.56} $^{\textcolor{red}{\uparrow 1.68}}$ & 84.03{\scriptsize$\pm$0.56} & 89.48{\scriptsize$\pm$0.58} $^{\textcolor{red}{\uparrow 1.45}}$ & 93.17{\scriptsize$\pm$0.70} & 78.31{\scriptsize$\pm$0.17} $^{\textcolor{red}{\uparrow 4.27}}$ & 83.15{\scriptsize$\pm$0.58} \\
            $\beta_{1}$-SLDC + ADE (ours) & {} & 87.18{\scriptsize$\pm$0.50} $^{\textcolor{red}{\uparrow 0.43}}$ & 92.42{\scriptsize$\pm$0.59} & 80.61{\scriptsize$\pm$0.36} $^{\textcolor{red}{\uparrow 4.99}}$ & 85.51{\scriptsize$\pm$0.41} & 91.36{\scriptsize$\pm$0.34} $^{\textcolor{red}{\uparrow 3.33}}$ & 94.37{\scriptsize$\pm$0.73} & 79.78{\scriptsize$\pm$0.24} $^{\textcolor{red}{\uparrow 5.74}}$ & 83.91{\scriptsize$\pm$0.40} \\
            $\beta_{2}$-SLDC + ADE (ours) & {} & 87.15{\scriptsize$\pm$0.50} $^{\textcolor{red}{\uparrow 0.40}}$ & 92.38{\scriptsize$\pm$0.57} & 80.50{\scriptsize$\pm$0.30} $^{\textcolor{red}{\uparrow 4.88}}$ & 85.45{\scriptsize$\pm$0.41} & 91.48{\scriptsize$\pm$0.24} $^{\textcolor{red}{\uparrow 3.45}}$ & 94.38{\scriptsize$\pm$0.69} & 80.00{\scriptsize$\pm$0.29} $^{\textcolor{red}{\uparrow 5.96}}$ & 84.01{\scriptsize$\pm$0.46} \\
			\bottomrule
		\end{tabular}
	}
\label{tab:results_sup21k}
\end{table*}

\subsubsection{Benchmarks.} To comprehensively evaluate the CIL performance, we conduct experiments on four widely-used benchmark datasets, i.e., CIFAR-100 \cite{krizhevsky2009learning}, ImageNet-R \cite{hendrycks2021many}, CUB-200 \cite{wah2011caltech}, and Cars-196 \cite{krause20133d}. Each dataset is uniformly partitioned into 10 disjoint tasks without any emphasis. The CIFAR-100 comprises 100 classes of natural images, with 500 training samples per class. The ImageNet-R contains images from 200 classes. Totally, it has 24,000 and 6,000 samples for training and test sets, respectively. Specifically, ImageNet-R is challenging for the PTMs because its images are either hard examples from ImageNet-21K or new images in diverse styles. CUB-200 contains 200 bird species with approximately 60 images per class. The training and test sets are split evenly. Cars-196 consists of 196 car types. It has 8,144 training and 8,040 testing images totally. Following the established protocols, CIFAR-100 and ImageNet-R serve as standard CIL benchmarks, while CUB-200 and Cars-196 evaluate fine-grained classification capabilities. All experiments are conducted using the PILOT framework \cite{sun2025pilot} with consistent random seeds to ensure fair comparison.
\subsubsection{Metrics.} We report two key metrics, i.e., the average classification accuracy across all classes encountered after each incremental task, denoted as Inc-Acc (\%), and the classification accuracy after completing the final task, denoted as Last-Acc (\%). The first metric evaluates the balance of remembering old classes and learning new ones throughout the CIL process, while the second one shows the overall performance across all classes after all tasks are learned. 
\subsubsection{CIL baselines.} 
Our proposed SLDC methods are compared against advanced PTM-based CIL approaches, including BiC \cite{Wu_2019_CVPR}, LwF \cite{li2017learning}, SLCA/SLCA++ \cite{Gao_2023_ICCV, zhang2024slcaunleashpowersequential}, RanPAC \cite{NEURIPS2023_2793dc35}, and CoMA/CoFiMA \cite{10.1007/978-3-031-73209-6_18}. Specifically, SeqKD denotes the distillation-enhanced SeqFT. Since $\alpha_{1,2}$-SLDC and $\beta_{1,2}$-SLDC methods are implemented based on SeqFT and SeqKD, respectively, the relative improvements over SeqFT and SeqKD are reported. Notably, our methods can be further integrated with other techniques, such as CoMA and CoFiMA, where EMA is employed on model parameters to mitigate representation drifts. As an upper-bound reference, the performance of joint training is reported, where the model is trained on all incremental tasks simultaneously. Additionally, MLPDC, which refers to the MLP-based distribution compensation method, also serves as a baseline method to SLDC-based compensation.

\subsubsection{Implementation details.} 
Two PTMs, which are the ViT-B/16 pre-trained on ImageNet-21K supervisedly \cite{ridnik2021imagenet} and the ViT-B/16 pre-trained using the MoCo-V3 self-supervised technique on ImageNet-1K \cite{Chen_2021_ICCV}, are employed. The LoRA adapters are of rank 4 and optimized using the Adam optimizer with a learning rate of $10^{-4}$ and a weight decay of $3 \times 10^{-5}$. For $\alpha_{1}$-SLDC, $\lambda_{\alpha_{1}}$ is set to $10^{-4}$. In the case of $\alpha_{2}$-SLDC, $\mathbf{A}_{t}$ and $\psi(f)$ are initialized as an identity matrix and a three-layer MLP with ReLU activation, respectively, where the hidden dimension of $\psi(f)$ matches that of the [cls] token in the ViTs. The default value for $\lambda_{\alpha_{2}}$ is 0.5, and the coefficients $(c_{1}, c_{2})$ are set to (0.9,0.1). Additional training details are provided in the Appendix. To re-estimate the class-specific mean and covariance in $\alpha_{2}$-SLDC through Gaussian sampling, we use $N = 10d$ samples per class, where $d$ denotes the feature dimension. For feature-based distillation in $\beta_{1,2}$-SLDC, let $\gamma_{\rm KD} = 1.0$ and $\gamma_{\rm Norm} = 0.1$ simply.

\subsection{Main comparison results}
Tables \ref{tab:results_mocov3} and \ref{tab:results_sup21k} present comprehensive comparisons between our proposed SLDC methods and state-of-the-art CIL approaches using both self-supervised (MoCo-V3) and supervised (ImageNet-21K) pre-trained ViT-B/16 backbones. In addition, the comparison results are visualized in Figs. \ref{fig:moco_results} and \ref{fig:sup21k_results} in the appendix. Some notable observations are as follows.
\begin{enumerate}
    \item Vanilla SeqFT struggles with severe forgetting, as evidenced by its low Last-Acc values, such as 64.40\% on CUB-200 and 61.37\% on ImageNet-R (see Table \ref{tab:results_mocov3}). In contrast, SLDC methods significantly boost accuracy without regularizing the backbone optimization. For example, $\alpha_2$-SLDC lifts CUB-200 performance to 78.98\% (a +14.58\% absolute gain) with MoCo-V3 architecture.
    
    \item When ADE is not employed, $\alpha_2$-SLDC consistently outperforms linear $\alpha_1$-SLDC and nonlinear MLPDC on fine-grained datasets, with notable gains on Cars-196 (77.53\% vs. 61.01\% with MoCo-V3) and CUB-200 (78.98\% vs. 70.42\%).    
    
    \item SeqKD improves SeqFT substantially, with a +12.57\% Last-Acc gain on CUB-200 using Sup-21K. Notably, distillation pairs exceptionally well with SLDC: $\beta_1$-SLDC (distillation-enhanced $\alpha_1$-SLDC) nearly matches $\alpha_2$-SLDC, such as 80.55\% vs. 78.98\% on CUB-200 with MoCo-V3.
    
    \item $\alpha_2$-SLDC and $\beta_2$-SLDC deliver robust performance across all datasets and pre-trained models. It outperforms MLPDC (nonlinear compensation) by +6.52\% on Cars-196 and +2.17\% on CIFAR-100 with Sup-21K pretraining, supporting our hypothesis that appropriate operators lie between linear and nonlinear extremes.
    
    \item ADE significantly enhances the performance of SLDC methods on fine-grained datasets. For example, $\alpha_1$-SLDC shows instability with Sup-21K pretraining, with Last-Acc dropping on CUB-200 (71.49\% vs. SeqFT’s 76.57\%) and Cars-196 (46.78\% vs. 54.53\%). Nonetheless, $\alpha_1$-SLDC + ADE achieves a striking +26.23\% improvement on Cars-196 compared to its non-ADE counterpart. This confirms ADE’s ability to mitigate approximation errors when task data is limited.
\end{enumerate}
\subsection{Ablation studies}
\subsubsection{Effectiveness to long-sequence CIL.}
\begin{figure*}[t]
    \centering
    \includegraphics[width=0.65\linewidth]{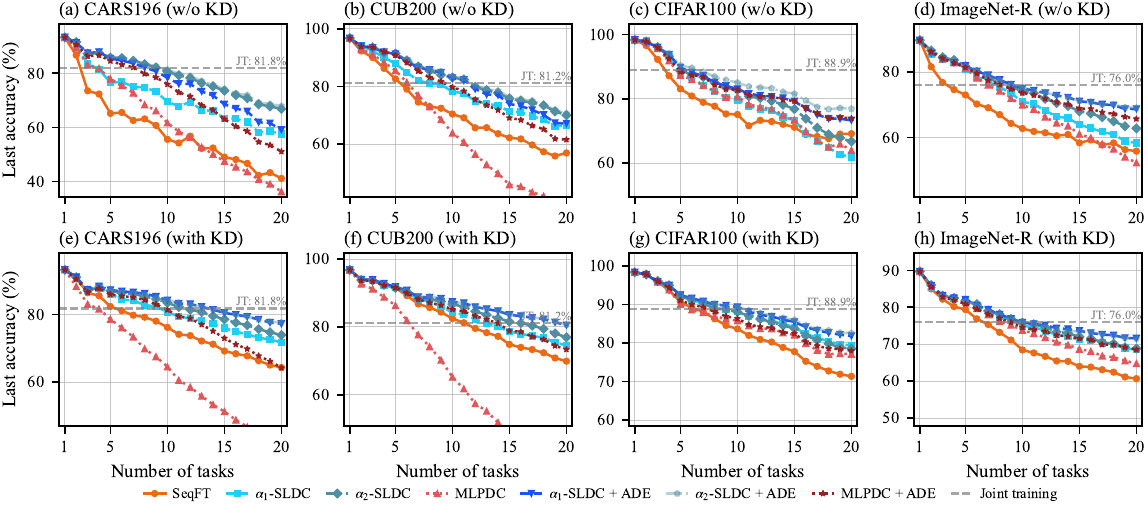}
    \caption{Performance comparison of SLDC methods on a 20-task sequence, demonstrating state-of-the-art results both with and without knowledge distillation.}
    \label{fig:sldc-long-sequence-moco}
\end{figure*}
Here, we extend the evaluation to 20 tasks to assess the effectiveness of SLDC methods on long-sequence CIL scenarios. The comparative results with and without distillation on the MoCo-V3 architecture are presented in Figure~\ref{fig:sldc-long-sequence-moco}, while the corresponding results for the Sup-21K architecture are provided in Figure~\ref{fig:sldc-long-sequence} in the Appendix. Some noteworthy observations are listed as follows. 1) The $\alpha_{2}$-SDLC approach consistently outperforms $\alpha_{1}$-SLDC when neither distillation nor ADE is applied. The incorporation of both distillation and ADE techniques yields significant improvements across all SLDC variants. 2) MLPDC exhibits particularly poor performance on the Cars196 and CUB200 datasets. 3) The $\alpha_{1}$-SLDC still suffers from instability when it is implemented on the Sup-21K architecture, and it can be mitigated effectively through either distillation or ADE techniques.
\subsubsection{Effectiveness to hybrid CIL datasets.}
\begin{figure}[t]
\centering
\includegraphics[width=0.85\linewidth]{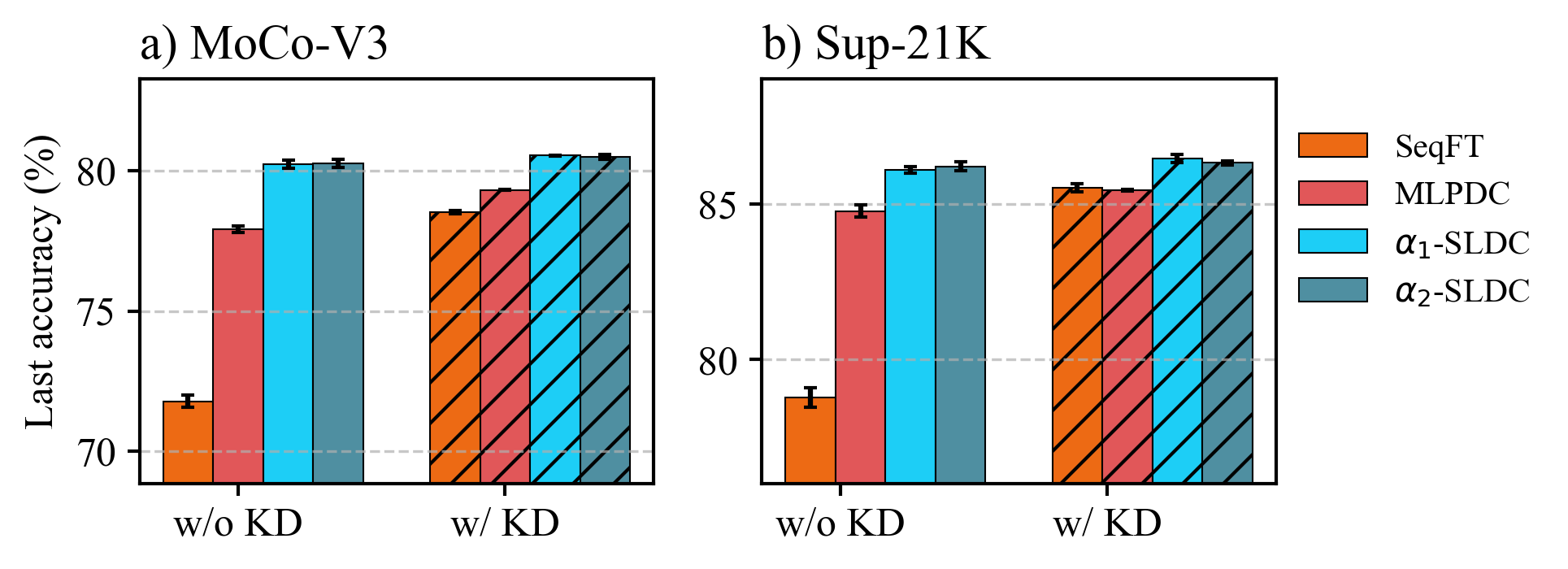}
\caption{Comparative performance of SLDC methods on hybrid CIL tasks comprising four distinct datasets: CIFAR-100, Cars-196, CUB-200, and ImageNet-R}
\label{fig:hybrid-performance}
\end{figure}
To evaluate the robustness of SLDC methods in heterogeneous CIL scenarios, we construct a hybrid CIL benchmark where each evaluation dataset (CIFAR-100, Cars-196, CUB-200, and ImageNet-R) is treated as a distinct incremental task. Figure~\ref{fig:hybrid-performance} presents the comparative results under both MoCo-V3 and Sup-21K pre-training strategies with and without distillation. Key findings include: 1) SLDC methods outperform both SeqFT and MLPDC baselines across all settings. 2) The performance gap between $\alpha_{1,2}$-SLDC methods narrows significantly in this setting. It means that $\alpha_1$-SLDC achieves comparable stability to its weak-nonlinear counterpart when dealing with larger task-specific datasets. In practice, we have tried experiments with varied dataset orders, and the evaluation results are similar. 
\subsubsection{Influences of $\alpha_{\rm temp}$ in $\alpha_{1}$-SLDC.}
\begin{figure}[h]
    \centering
    \includegraphics[width=0.75\linewidth]{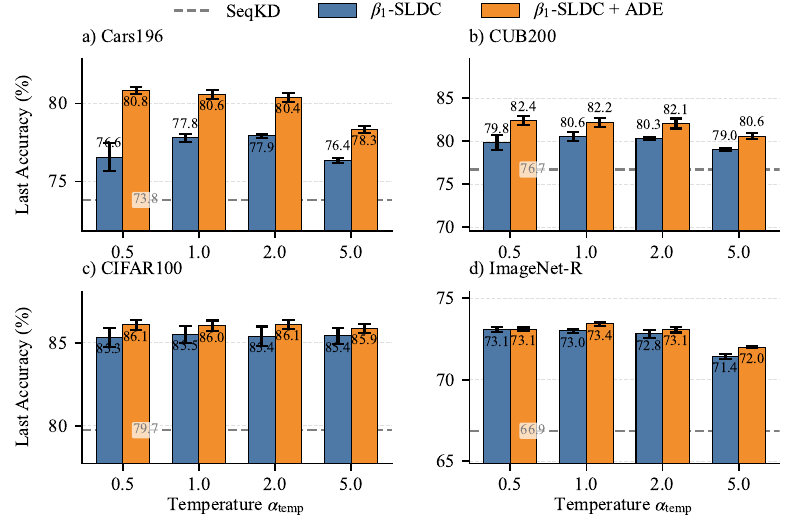}
    \caption{Performance comparison with varying temperature parameters $\alpha_{\rm temp}$ in $\alpha_{1}$-SLDC}
    \label{fig:influence_alpha_temp}
\end{figure}
\begin{figure}[h]
    \centering   
    \includegraphics[width=0.75\linewidth]{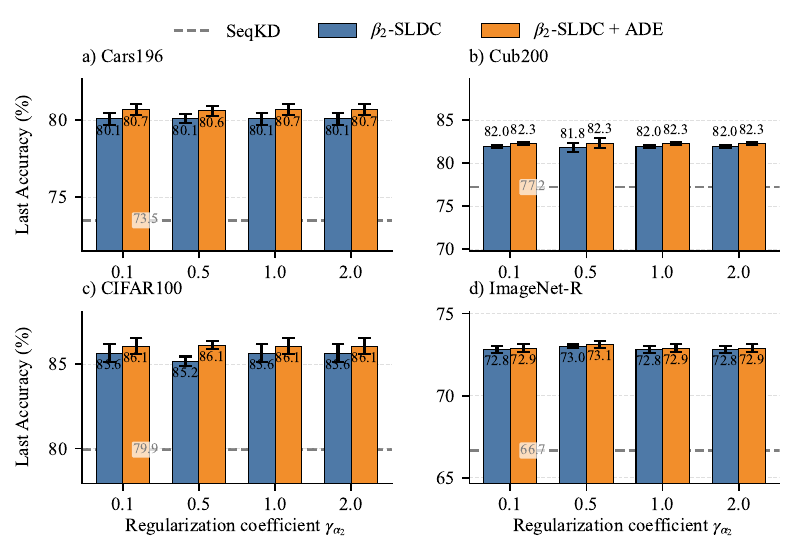}
    \caption{Performance evaluation of $\alpha_{2}$-SLDC with varying regularization coefficients $\gamma_{\alpha_{2}} \in \{0.1, 0.5, 1.0, 2.0\}$}
    \label{fig:gamma_two_test}
\end{figure}
\begin{figure}[b]
    \centering
    \includegraphics[width=0.75\linewidth]{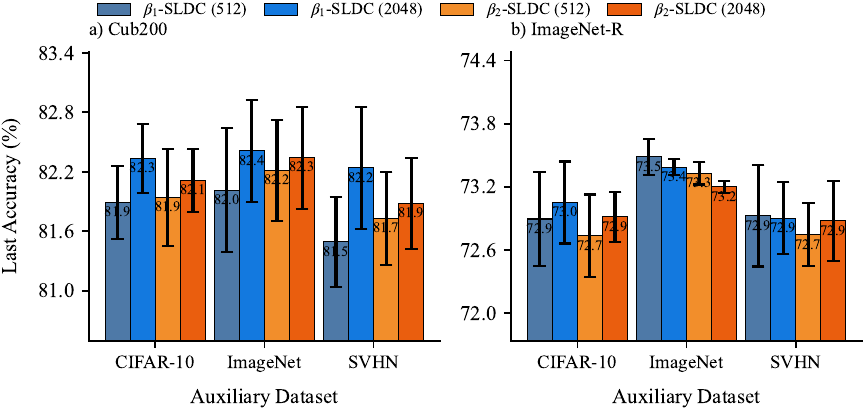}
    \caption{Performance comparison of SLDC methods with varying ADE datasets (CIFAR-10, SVHN, and ImageNet) and sample sizes (512 to 2048)}
    \label{fig:sensitivity_analysis_on_ADE_datasets}
\end{figure}
This part analyzes the impact of the temperature parameter $\alpha_{\rm temp}$ in $\alpha_1$-SLDC. Focusing on the MoCo-V3 architecture with distillation, we evaluate four $\alpha_{\rm temp}$ values ($[0.5, 1.0, 2.0, 5.0]$), with results shown in Figure~\ref{fig:influence_alpha_temp}. Our experiments reveal two key findings. 1) When ADE is not employed, $\alpha_{\rm temp}=1.0$ achieves optimal performance on fine-grained datasets Cars-196 and CUB-200. (2) When ADE is employed, reducing $\alpha_{\rm temp}$ below 1.0 becomes advantageous for effectively utilizing the unlabeled dataset. These findings suggest that the optimal temperature depends on whether ADE is implemented.

\subsubsection{Influences of $\gamma_{\alpha_{2}}$. } 
Here, we investigate the influences of regularization coefficient $\gamma_{\alpha_{2}}$ in $\alpha_{2}$-SLDC by selecting values from $[0.1, 0.5, 1.0, 2.0]$. For simplicity, the results on the MoCo-V3 architecture with distillation are reported in Figure~\ref{fig:gamma_two_test}. The performance of $\alpha_{2}$-SLDC exhibits remarkable stability across the tested range of $\gamma_{\alpha_{2}}$ values. It suggests that the prior assumption governing the hypothesis space of the transition operator plays a more critical role than the specific choice of the regularization coefficient.

\subsubsection{Sensitivity to sample selection in ADE.}
This section examines the impact of sample selection in the ADE process. We evaluate three ADE datasets (CIFAR-10, SVHN, and ImageNet) with varying sample sizes ranging from 512 to 2048. As shown in Figure~\ref{fig:sensitivity_analysis_on_ADE_datasets}, our analysis reveals distinct patterns across different benchmark datasets. For the fine-grained CUB-200 dataset, all ADE variants improve SLDC performance, with larger ADE sample sizes yielding progressively better results. In contrast, the ImageNet-R dataset demonstrates stable performance without requiring ADE, suggesting that the training samples in ImageNet-R is sufficient to achieve robust performance.

\section{Conclusions}
In this paper, an in-depth exploration on pre-trained ViT-based CIL is conducted, and it is highlighted that effective approximation of the latent space transition operator is critical for mitigating the adverse effects of distribution drifts during sequential optimization. Accordingly, the linear $\alpha_1$-SLDC and weak-nonlinear $\alpha_2$-SLDC methods are introduced, along with their distillation-enhanced variants, $\beta_1$-SLDC and $\beta_2$-SLDC, to align the distributions of previous classes with the updated feature space. 
Extensive experiments demonstrate the efficacy of our methods. Notably, the synergy of distillation (to limit excessive optimization) and SLDC (to compensate for distribution drifts) significantly narrows the performance gap between CIL and joint learning, making CIL more practical for real-world applications.

However, we observed that $\alpha_1$-SLDC exhibits instability on certain fine-grained datasets with the Sup-21K architecture, and auxiliary unlabeled data are required to stabilize its performance. In addition, the applicability of SLDC methods to multi-modal models remains an open question, which we plan to explore in our future works.

\section{Acknowledgment}
This work was supported in part by the National Natural Science Foundation of China under Grant 62573062, in part by the Shenzhen Science and Technology Program under grant JCYJ20230807093513027, and in part by the Fundamental Research Funds for the Central Universities under grant 1243300008.

\bibliography{aaai2026}

\newpage

\newpage
\section{Appendix}
\subsubsection{Algorithm}
The pseudocode of SLDC methods is summarized in Algorithm \ref{alg:sldc}. 
\begin{algorithm*}
\caption{Sequential Learning with Drift Compensation (SLDC) for Pre-trained ViT-based CIL}
\label{alg:sldc}
\begin{algorithmic}[1]
\REQUIRE  Pre-trained ViT $\mathcal{F}_{\theta}$, Initial classifier $\mathcal{C}_\phi$, Training datasets $\{\mathcal{D}_t\}_{t=1}^T$ \ 
    Hyperparameters:$ \gamma_{\text{KD}}, \gamma_{\text{Norm}}, \gamma_{\alpha_1}, \gamma_{\alpha_2}$

\STATE Initialize Gaussian distribution set: $\mathcal{H}_0 \gets \emptyset$. Initialize the cumulative set of observed classes: $\mathcal{C}_{t-1} = \emptyset$. 
\FOR{each task $t = 1$ \TO $T$}
    \STATE \textcolor{red}{\textbf{// Phase 1: Sequential Model Adaptation}}
    \STATE Expand classifier $\mathcal{C}_\phi \gets \text{Linear}(\mathbb{R}^d, |\mathcal{C}_{t-1} \cup \mathcal{Y}_t|)$
    \STATE Update observed classes $\mathcal{C}_{t} \leftarrow \mathcal{C}_{t-1} \cup \mathcal{Y}_{t}$
    \STATE Save backbone checkpoint $\theta_{t-1} \gets \theta$
    \FOR{each epoch $= 1$ \TO $N_{\text{epochs}}$}
        \FOR{each batch $(x,y) \in \mathcal{D}_t$}
            \STATE Extract features: $f \gets \mathcal{F}_{\theta}(x)$
            \STATE Compute classification loss: $\mathcal{L}_{\text{CE}} \gets \text{CE}(\mathcal{C}_\phi(f), y)$
            \STATE $\mathcal{L} \gets \mathcal{L}_{\text{CE}}$
            \IF{using distillation}
                \STATE $f_{\text{prev}} \gets \mathcal{F}_{\theta_{t-1}}(x)$, \ $\mathcal{L}_{\text{KD}} \gets \gamma_{\text{KD}} \cdot \|f_{\text{prev}} - f\|^2_2$, \ $\mathcal{L}_{\text{Norm}} \gets \gamma_{\text{Norm}} \cdot (\|f_{\text{prev}}\|_2 - \|f\|_2)^2$
            \STATE $\mathcal{L} \gets \mathcal{L} + \mathcal{L}_{\text{KD}} + \mathcal{L}_{\text{Norm}}$
            \ENDIF
            \STATE Update $\theta, \phi$ via $\nabla_{\theta,\phi}\mathcal{L}$
        \ENDFOR
    \ENDFOR
   \STATE \textcolor{blue}{\textbf{// Phase 2: Distribution Compensation}}
    \STATE Compute $F_{\mathcal{Y}_t}^{t-1} = \left[ \mathcal{F}_{\theta_{t-1}}(x_{t,1}), \dots, \mathcal{F}_{\theta_{t-1}}(x_{t,n_t}) \right] \in \mathbb{R}^{d \times n_t}$ and $F_{\mathcal{Y}_t}^{t} = \left[ \mathcal{F}_{\theta_{t}}(x_{t,1}), \dots, \mathcal{F}_{\theta_{t}}(x_{t,n_t}) \right] \in \mathbb{R}^{d \times n_t}$ 
    \STATE Calculate $\tilde{F}^{t-1}$ and $\tilde{F}^{t}$ by normalizing the elements of $F_{\mathcal{Y}_t}^{t-1}$ and $F_{\mathcal{Y}_t}^{t}$ to unit vectors. 
    \IF{use $\alpha_1$-SLDC}
        \STATE Compute the linear transformation matrix:
        $\mathbf{A}_t \gets \tilde{F}^{t} (\tilde{F}^{t-1})^\top (\tilde{F}^{t-1} (\tilde{F}^{t-1})^\top + \gamma_{\alpha_1} I)^{-1}$
        \STATE Apply sample complexity-based re-weighting: $w \gets \exp(-|\mathcal{D}_t| / d)$; $\mathbf{A}_t \gets (1-w) \mathbf{A}_t + w I$
        \FOR{each $(\mu_c, \Sigma_c) \in \mathcal{H}_{t-1}$}
            \STATE $\mu_c \gets \mathbf{A}_t \mu_c$
            \STATE $\Sigma_c \gets \mathbf{A}_t \Sigma_c \mathbf{A}_t^\top$
        \ENDFOR
    \ELSIF{use $\alpha_2$-SLDC}
        \STATE Initialize $\mathbf{A} \gets I_d$, $\psi \gets \text{MLP}(d \to h \to d)$, $(c_{1},c_{2}) \gets (0.9,0.1)$
\STATE $\min_{\mathbf{A}, \psi, \alpha_{1/2}} \| c_1 \mathbf{A}\tilde{F}^{t-1} + c_2 \psi(\tilde{F}^{t-1}) - \tilde{F}^{t}\|_F^2 + \gamma_{\alpha_2} (c_1  - 1)^2$
        \STATE Monte Carlo transformation:
        \FOR{each $(\mu_c, \Sigma_c) \in \mathcal{H}_{t-1}$}
            \STATE Sample $\{f_c^{(i)}\}_{i=1}^N \sim \mathcal{N}(\mu_c, \Sigma_c)$
            \STATE Transform $\tilde{f}_c^{(i)} \gets \alpha_1 \mathbf{A} f_c^{(i)} + \alpha_2 \psi(f_c^{(i)})$
            \STATE Re-estimate $\mu_c \gets \text{mean}(\{\tilde{f}_c^{(i)}\})$, $\Sigma_c \gets \text{cov}(\{\tilde{f}_c^{(i)}\})$
        \ENDFOR
    \ENDIF

    \STATE \textcolor{green!70!black}{\textbf{// Phase 3: Classifier Refinement}}
    \STATE Update distribution set: $\mathcal{H}_t \gets \mathcal{H}_{t-1} \cup \{ \text{New Gaussians for } \mathcal{Y}_t \}$
    \STATE Generate synthetic features: $\mathcal{F}_{\text{synth}} \gets \bigcup_{c} \{ f \sim \mathcal{N}(\mu_c, \Sigma_c) \}_{c \in \mathcal{C}_t}$
    \STATE Refine classifier: 
    $\phi \gets \arg \min_{\phi} \mathbb{E}_{(f,c) \sim \mathcal{F}_{\text{synth}}} [-\log p_\phi(c|f)]$
\ENDFOR
\end{algorithmic}
\end{algorithm*}

\subsection{Implementation details}
\subsubsection{LoRA settings.}
For a pre-trained weight matrix $W \in \mathbb{R}^{d_2 \times d_1}$, LoRA introduces a low-rank updating $\Delta W = BA$, where $A \in \mathbb{R}^{k \times d_1}$ and $B \in \mathbb{R}^{d_2 \times k}$, with rank $k \ll \min(d_1, d_2)$. During training, only the low-rank matrices $A$ and $B$ are updated, while the original weight matrix $W$ remains frozen.

Following the settings from SLCA++ \cite{zhang2024slcaunleashpowersequential}, we leverage singular value decomposition (SVD) to enhance the initialization of LoRA adapters. Specifically, for a pre-trained weight matrix $W$, SVD decomposes it as $W = U \Sigma_s V^\top$, where $U \in \mathbb{R}^{d_2 \times d_2}$ and $V \in \mathbb{R}^{d_1 \times d_1}$ are the left and right singular vectors, respectively, and $\Sigma_s \in \mathbb{R}^{d_2 \times d_1}$ contains the singular values. Then, $A$ is initialized by the top-$k$ rows of $V^\top$, while $B$ is initialized by zero values. By SVD-based initialization, it can be ensured that $BA = 0$ and the learning subspace of $A$ is aligned with the principal directions of $W$ at the start of training.

In all experiments, we let $k = 4$ and apply LoRA adapters to tune both the attention and MLP blocks. The parameter comparison between the full fine-tuning and the LoRA-based tuning are summarized in Table \ref{tab:22}.
\begin{table}[h]
\centering
\caption{Comparison of trainable parameters between the full fine-tuning and LoRA-based tuning}
\begin{tabular}{lrr}
\toprule
\textbf{Parameter Type} & \textbf{Count} & \textbf{Percentage of Total} \\
\midrule
Total parameters & 86,314,752 & 100.00\% \\
LoRA parameters & 516,096 & 0.60\% \\
\bottomrule
\end{tabular}
\label{tab:22}
\end{table}

\subsubsection{Backbone optimization.}
The training process employs 15 epochs for the CUB-200 and Cars-196 datasets, 5 epochs for CIFAR-100, and 10 epochs for ImageNet-R. All other hyperparameters remain consistent across the four datasets. For ViT backbones, an Adam optimizer is employed with an initial learning rate of $10^{-4}$, which is reduced to $\frac{1}{3} \times 10^{-4}$ in the final epoch. The linear classifier uses a learning rate which is 10 times higher than that of the ViT backbones.
\subsubsection{Optimization of weak-nonlinear and MLP transformations in $\alpha_{2}$-SLDC and MLPDC.}
To optimize the weak-nonlinear transformation $\psi(f)$ in $\alpha_{2}$-SLDC and the MLP in MLPDC, the following configurations are employed. 
\begin{enumerate}
    \item Both the weak-nonlinear and MLP transformations are optimized for 5,000 steps totally. The batch size is 32. 
    \item The Adam optimizer is employed with an initial learning rate of $10^{-3}$, which is reduced to $5 \times 10^{-4}$ at the last optimization step.
    \item The models take normalized pre-optimization features $\tilde{F}_{\mathcal{Y}_{t}}^{t-1}$ as inputs and are trained to minimize the mean squared error (MSE) loss between the predictive values and the normalized post-optimization features $\tilde{F}_{\mathcal{Y}_{t}}^{t}$.
    \item For the weak-nonlinear transformation $\psi(f)$, no weight decay is employed. In contrast, for MLP transformation, a weight decay of $10^{-6}$ is applied to mitigate overfitting.
\end{enumerate}
\subsection{Introduction to benchmark datasets}
Below is a detailed introduction to the four benchmark datasets used for evaluating CIL performance. These datasets are widely adopted in the machine learning community for their diversity and ability to test various aspects of model performance, particularly in CIL and fine-grained classification tasks.

\subsubsection{CIFAR-100}
CIFAR-100 is a standard benchmark for image classification and CIL tasks. It comprises 100 classes of natural images, which are grouped into 20 superclasses (e.g., vehicles, animals, household items). The dataset contains 60,000 color images of size 32$\times$32 pixels, with 500 training samples and 100 test samples per class, resulting in 50,000 training and 10,000 test images. In particular, we resize the resolution to $224\times224$ in our experiments. 

\subsubsection{ImageNet-R}
ImageNet-R (R for Renditions) includes 200 classes, featuring hard examples from ImageNet-21K and new images in diverse styles, such as cartoons, paintings, and sketches. The dataset consists of 30,000 images in total, with 24,000 training and 6,000 test images. The diversity in visual styles and the inclusion of difficult examples make ImageNet-R a rigorous benchmark for assessing generalization in incremental learning scenarios.

\subsubsection{CUB-200}
The CUB-200 dataset is tailored for fine-grained classification. It includes 200 distinct bird species. The dataset contains approximately 11,788 high-resolution images. Each class has nearly 60 images on average. The dataset is evenly divided into training and test sets. Each set includes roughly 30 images per class. 

\subsubsection{Cars-196}
The Cars-196 dataset serves as an another fine-grained classification benchmark. It comprises 196 distinct car types. The dataset includes 16,184 high-resolution images, which are split into 8,144 training images and 8,040 test images. It captures subtle differences in car designs, such as headlights, grilles, or body
shapes across various angles and lighting conditions.
\begin{table*}[htbp]
\centering
\caption{Dataset descriptions}
\begin{tabular}{lcccc}
\toprule
Dataset & Classes & Train Images & Test Images & Resolution \\
\midrule
ImageNet-R & 200 & 24,000 & 6,000 & 224$\times$224 \\
CUB200& 200 & 5,994 & 5,794 & 224$\times$224 \\
CIFAR100 & 100 & 50,000 & 10,000 & 224$\times$224 \\
Cars196 & 196 & 8,054 & 8,131 & 224$\times$224 \\
\bottomrule
\end{tabular}
\label{tab:datasets}
\end{table*}
\subsection{Introduction to implemented pre-trained ViT}
This paper employes two pre-trained models, i.e., the MoCo-V3 ViT-B/16 and Sup21K ViT-B/16. Detailed descriptions of these models are provided below.
\subsubsection{MoCo-V3 architecture.} MoCo-V3 is a self-supervised learning framework which extends the momentum contrast (MoCo) approach to ViTs. The MoCo approach employs a query encoder and a momentum-updated key encoder, along with a contrastive loss, to learn robust visual representations without labeled data. In this study, the ViT-B/16 backbone pre-trained with MoCo-V3 on ImageNet-1K is employed.
\subsubsection{Sup-21K architecture.} The Sup-21K model refers to the ViT-B/16 architecture pre-trained in a fully supervised manner on the large-scale ImageNet-21K dataset. This model benefits from rich semantic supervision across 21K categories, providing robust and transferable feature representations for downstream tasks.
\subsection{More analytical results}
\begin{statement}[Solution to the regularized least-squares problem]
The regularized least-squares problem for estimating the linear operator \(\mathbf{A}_t\) is formulated as
\[
\mathbf{A}_t = \arg\min_{\mathbf{A}} \left\| \mathbf{A} \tilde{F}_{\mathcal{Y}_t}^{t-1} - \tilde{F}_{\mathcal{Y}_t}^{t} \right\|_F^2 + \lambda \left\| \mathbf{A} \right\|_F^2,
\]
where \(\tilde{F}_{\mathcal{Y}_t}^{t-1} \in \mathbb{R}^{d \times n_t}\) and \(\tilde{F}_{\mathcal{Y}_t}^{t} \in \mathbb{R}^{d \times n_t}\) are column-wise \(L_2\)-normalized feature matrices, \(\lambda = \gamma_{\alpha_1} > 0\) is the regularization coefficient, and \(\|\cdot\|_F\) denotes the Frobenius norm. The analytical solution to this problem is given by
\[
\mathbf{A}_t = \tilde{F}_{\mathcal{Y}_t}^{t} (\tilde{F}_{\mathcal{Y}_t}^{t-1})^\top \left( \tilde{F}_{\mathcal{Y}_t}^{t-1} (\tilde{F}_{\mathcal{Y}_t}^{t-1})^\top + \lambda I_d \right)^{-1},
\]
where \(I_d\) is the \(d \times d\) identity matrix.
\label{theorm:LS}

\end{statement}
\begin{proof}
Let \(X = \tilde{F}_{\mathcal{Y}_t}^{t-1}\) and \(Y = \tilde{F}_{\mathcal{Y}_t}^{t}\). The optimization objective is
\begin{align}
\min_{\mathbf{A}} J(\mathbf{A}) = \| \mathbf{A} X - Y \|_F^2 + \lambda \| \mathbf{A} \|_F^2.
\end{align}
By expressing the Frobenius norms as traces, we obtain
\begin{align}
\| \mathbf{A} X - Y \|_F^2 &= \operatorname{tr}(X^\top \mathbf{A}^\top \mathbf{A} X) - 2\operatorname{tr}(X^\top \mathbf{A}^\top Y) \nonumber \\ 
&\quad + \operatorname{tr}(Y^\top Y),
\label{eq::ff}
\end{align}
and 
\begin{align}
\lambda \| \mathbf{A} \|_F^2 &= \lambda \operatorname{tr}(\mathbf{A}^\top \mathbf{A}).
\label{eq:trace}
\end{align}
By combining \eqref{eq::ff} with \eqref{eq:trace}, we obtain
\begin{align}
J(\mathbf{A}) &= \operatorname{tr}(X^\top \mathbf{A}^\top \mathbf{A} X) - 2\operatorname{tr}(X^\top \mathbf{A}^\top Y) \nonumber \\
&\quad + \operatorname{tr}(Y^\top Y) + \lambda \operatorname{tr}(\mathbf{A}^\top \mathbf{A}).
\end{align}
Taking the partial derivative of \(J(\mathbf{A})\) with respect to \(\mathbf{A}\) and setting it to zero, we have
\begin{align}
\frac{\partial J}{\partial \mathbf{A}} = 2\mathbf{A} X X^\top - 2 Y X^\top + 2\lambda \mathbf{A} = 0,
\end{align}
which can be simplified to
\begin{align}
\mathbf{A} (X X^\top + \lambda I_d) = Y X^\top.
\end{align}
Since \(X X^\top + \lambda I_d\) is invertible for \(\lambda > 0\), the solution is obtained by
\begin{align}
\mathbf{A} = Y X^\top (X X^\top + \lambda I_d)^{-1}.
\label{eq:ls_solution}
\end{align}
By substituting \(X\) and \(Y\) into the \eqref{eq:ls_solution}, we obtain the closed-form solution to \(\mathbf{A}_t\).
\end{proof}

\begin{statement}[Linear transformation of a Gaussian distribution]
    Let \(\mathbf{x} \in \mathbb{R}^d\) be a random vector following a Gaussian distribution, \(\mathbf{x} \sim \mathcal{N}(\boldsymbol{\mu}, \boldsymbol{\Sigma})\). For any invertible linear transformation \(\mathbf{y} = \mathbf{A} \mathbf{x}\), where \(\mathbf{A} \in \mathbb{R}^{d \times d}\) is an invertible matrix, the random vector \(\mathbf{y}\) follows a Gaussian distribution, \(\mathbf{y} \sim \mathcal{N}(\mathbf{A} \boldsymbol{\mu}, \mathbf{A} \boldsymbol{\Sigma} \mathbf{A}^\top)\).
    \label{theorm:linear_transformation}
\end{statement}
\begin{proof}
 The mean of \(\mathbf{y}\) is given by
\begin{align}
\mathbb{E}[\mathbf{y}] = \mathbb{E}[\mathbf{A} \mathbf{x}] = \mathbf{A} \mathbb{E}[\mathbf{x}] = \mathbf{A} \boldsymbol{\mu}.
\end{align}
The covariance of \(\mathbf{y}\) is computed by
\begin{align}
\text{Cov}(\mathbf{y}) &= \mathbb{E}\left[(\mathbf{y} - \mathbb{E}[\mathbf{y}])(\mathbf{y} - \mathbb{E}[\mathbf{y}])^\top\right] \nonumber \\
&= \mathbb{E}\left[(\mathbf{A} \mathbf{x} - \mathbf{A} \boldsymbol{\mu})(\mathbf{A} \mathbf{x} - \mathbf{A} \boldsymbol{\mu})^\top\right] \nonumber \\
&= \mathbb{E}\left[\mathbf{A} (\mathbf{x} - \boldsymbol{\mu})(\mathbf{x} - \boldsymbol{\mu})^\top \mathbf{A}^\top\right] \nonumber \\
&= \mathbf{A} \mathbb{E}\left[(\mathbf{x} - \boldsymbol{\mu})(\mathbf{x} - \boldsymbol{\mu})^\top\right] \mathbf{A}^\top \nonumber \\
&= \mathbf{A} \boldsymbol{\Sigma} \mathbf{A}^\top.
\end{align}
Since \(\mathbf{x}\) is Gaussian and \(\mathbf{A}\) is invertible, \(\mathbf{y} = \mathbf{A} \mathbf{x}\) is also Gaussian (as linear transformations preserve Gaussianity). Thus, \(\mathbf{y} \sim \mathcal{N}(\mathbf{A} \boldsymbol{\mu}, \mathbf{A} \boldsymbol{\Sigma} \mathbf{A}^\top)\).
\end{proof}

\begin{statement}[The affinity of transition operator under the NTK limits]
Let \( f_{\theta_0}: \mathcal{X} \to \mathbb{R}^{d} \) be a pre-trained neural network with pre-trained parameters \( \theta_0 \). During the fine-tuning, it is trained on a dataset \( \{(x_i, y_i)\}_{i=1}^n \) where \( y_i \in \mathbb{R}^d \). The loss function is the mean squared error
\begin{align}
\mathcal{L}(\theta) = \frac{1}{2} \sum_{i=1}^n \| f_\theta(x_i) - y_i \|^2.
\end{align}
Assuming that the network width \( d \to \infty \), by the infinite NTK theory, the NTK
\begin{align}
\Theta_{\theta_0}(x, x') = \left[ \nabla_\theta f_{\theta_0}(x) \right]^\top \left[ \nabla_\theta f_{\theta_0}(x') \right] \in \mathbb{R}^{d \times d}
\end{align}
is deterministic and constant during the fine-tuning. Suppose the learning rate $\eta$ satisfies \(\eta = O(1/\| \Theta_{\theta_0} \|_{\rm op})\), where \(\|\cdot\|_{\rm op}\) is the operator norm of the NTK Gram matrix. After one gradient descent step, we have
\begin{align}
    \theta_1 = \theta_0 - \eta \nabla_\theta \mathcal{L}(\theta_0).
\end{align}
Specifically, the updated function satisfies the following affinity form
\begin{align}
f_{\theta_1}(x) = A[f_{\theta_0}](x) + b(x), \quad \forall x \in \mathcal{X},
\end{align}
where
\( A: \mathcal{F} \to \mathcal{F} \) is the linear operator
\begin{align}
    A[f](x) = f(x) - \eta \sum_{i=1}^n \Theta_{\theta_0}(x, x_i) f(x_i),
\end{align}
and \( b(x) = \eta \sum_{i=1}^n \Theta_{\theta_0}(x, x_i) y_i \) is the input-dependent bias function.
\label{state:affinity}
\end{statement}
\begin{proof}
The gradient of \( \mathcal{L} \) at \( \theta_0 \) is
\begin{align}
\nabla_\theta \mathcal{L}(\theta_0) = \sum_{i=1}^n \left[ \nabla_\theta f_{\theta_0}(x_i) \right] (f_{\theta_0}(x_i) - y_i).
\end{align}
Let \( J_i = \nabla_\theta f_{\theta_0}(x_i) \in \mathbb{R}^{p \times d} \) be the Jacobian matrix and \( r_i = f_{\theta_0}(x_i) - y_i \in \mathbb{R}^d \) be the residual terms. Then, we have
\begin{align}
\nabla_\theta \mathcal{L}(\theta_0) = \sum_{i=1}^n J_i r_i.
\label{eq:update_grad}
\end{align}
The parameter updating is 
\begin{align}
\theta_1 - \theta_0 = -\eta \sum_{i=1}^n J_i r_i.
\end{align}
For any \( x \in \mathcal{X} \), with the first-order Taylor expansion of $f(x)$, we have 
\begin{align}
f_{\theta_1}(x) - f_{\theta_0}(x) =  J_x^\top (\theta_1 - \theta_0) + O(\|\theta_1 - \theta_0\|^2),
\label{eq:Taylor}
\end{align}
where \( J_x = \nabla_\theta f_{\theta_0}(x) \). Substituting \eqref{eq:update_grad} into the \eqref{eq:Taylor}, we obtain
\begin{align}
f_{\theta_1}(x) - f_{\theta_0}(x) = -\eta \sum_{j=1}^n J_x^\top J_j r_j + O(\eta^2).
\label{eq:xxxx}
\end{align}
By the NTK definition \( J_x^\top J_j = \Theta_{\theta_0}(x, x_j) \), \eqref{eq:xxxx} can be formulated by
\begin{align}
f_{\theta_1}(x) - f_{\theta_0}(x) = -\eta \sum_{j=1}^n \Theta_{\theta_0}(x, x_j) r_j + O(\eta^2),
\end{align}
which can be further reformulated by
\begin{align}
    & f_{\theta_1}(x) - f_{\theta_0}(x)  = -\eta \sum_{j=1}^n \Theta_{\theta_0}(x, x_j) f_{\theta_0}(x_j) \nonumber \\ &+ \eta \sum_{j=1}^n \Theta_{\theta_0}(x, x_j) y_j + O(\eta^2).
\end{align}
As \( d \to \infty \), \( O(\eta^2) \to 0 \), we have
\begin{align}
& f_{\theta_1}(x) = f_{\theta_0}(x) \nonumber \\& - \eta \sum_{i=1}^n \Theta_{\theta_0}(x, x_i) f_{\theta_0}(x_i) + \eta \sum_{i=1}^n \Theta_{\theta_0}(x, x_i) y_i.
\end{align}
This yields the affinity form \( f_{\theta_1}(x) = A[f_{\theta_0}](x) + b(x) \).
To demonstrate the linearity of $A$, for any \( f, g \in \mathcal{F} \) and \( \alpha, \beta \in \mathbb{R} \), we have 
\begin{align}
& A[\alpha f + \beta g](x) \nonumber \\ &= (\alpha f(x) + \beta g(x)) - \eta \sum_{i=1}^n \Theta_{\theta_0}(x, x_i) (\alpha f(x_i) + \beta g(x_i)) \nonumber  \\
&= \alpha \left( f(x) - \eta \sum_{i=1}^n \Theta_{\theta_0}(x, x_i) f(x_i) \right) + \nonumber  \\ & \ \ \ \ \ \beta \left( g(x) - \eta \sum_{i=1}^n \Theta_{\theta_0}(x, x_i) g(x_i) \right) \nonumber  \\
&= \alpha A[f](x) + \beta A[g](x).
\end{align}
Thus, \( A \) is a linear operator on \( \mathcal{F} \).
\end{proof}

\begin{remark}
    In particular, the statement \ref{state:affinity} claims that the updated function after one gradient descent step takes the affine form \( f_{\theta_1}(x) = A[f_{\theta_0}](x) + b(x) \). However, the operator \( A \) is not equivalent to multiplication by a real-valued matrix \( P \in \mathbb{R}^{d \times d} \), i.e., \( f_{\theta_1}(x) \neq P f_{\theta_0}(x) + b(x) \). The reason is as follows. The operator \( A \) is an integral-type operator (specifically, a discrete sum approximating an integral) that depends on the entire training set. It maps a function \( f \) to a new function \( A[f] \) by combining pointwise evaluation at \( x \) with a weighted sum of evaluations at all training points \( x_i \). The weights \( \Theta_{\theta_0}(x, x_i) \in \mathbb{R}^{d \times d} \) are matrix-valued and vary with both \( x \) and \( x_i \). This makes \( A \) a global operator that cannot be reduced to a pointwise matrix multiplication. Nonetheless, restricting \( x \) to the training set allows an equivalent representation using a real-valued matrix \( P \). Below, we state and prove this as a new theorem. 

\end{remark}
\begin{statement}
    Consider the training set \( S = \{x_1, \dots, x_n\} \). Define the following notations: 
    \begin{enumerate}
        \item The vector of function values: \( \mathbf{f}_{\theta} = \begin{bmatrix} f_{\theta}(x_1) \\ \vdots \\ f_{\theta}(x_n) \end{bmatrix} \in \mathbb{R}^{n  d} \).
        \item The vector of labels: \( \mathbf{y} = \begin{bmatrix} y_1 \\ \vdots \\ y_n \end{bmatrix} \in \mathbb{R}^{n \cdot d} \).
        \item The NTK Gram matrix \( \mathbf{\Theta} \in \mathbb{R}^{(n  d) \times (n  d)} \), which is a block matrix where the \((k,j)\)-th block is \( \Theta_{\theta_0}(x_k, x_j) \in \mathbb{R}^{d \times d} \).

    \end{enumerate}
After one gradient descent step with learning rate \( \eta \), the updated function values on the training points satisfy:
\[
\mathbf{f}_{\theta_1} = P \mathbf{f}_{\theta_0} + \mathbf{b},
\]
where \( P = I - \eta \mathbf{\Theta} \) is a real-valued matrix, \( \mathbf{b} = \eta \mathbf{\Theta} \mathbf{y} \), and \( I \) is the identity matrix with dimension \( n d \).
\label{theorem:training}
\end{statement}
 \begin{proof}
From the proof of Statement \ref{state:affinity}, for any \( x \in \mathcal{X} \), the first-order Taylor expansion of $f(x)$ in the infinite-width limit (\( d \to \infty \)) follows 
\begin{align}
 f_{\theta_1}(x)  & = f_{\theta_0}(x) -  \eta \sum_{j=1}^n \Theta_{\theta_0}(x, x_j) f_{\theta_0}(x_j) \nonumber \\ & \quad + \ \eta \sum_{j=1}^n \Theta_{\theta_0}(x, x_j) y_j.
\end{align}
Hereafter, by evaluating it at a training point \( x_k \) (where \( k \in \{1, \dots, n\} \)), we get 
\begin{align}
    f_{\theta_1}(x_k) = & \ f_{\theta_0}(x_k) - \eta \sum_{j=1}^n \Theta_{\theta_0}(x_k, x_j) f_{\theta_0}(x_j) \nonumber \\ & + \eta \sum_{j=1}^n \Theta_{\theta_0}(x_k, x_j) y_j.
\end{align}
Define the vector \( \mathbf{f}_{\theta_1} \) by stacking \( f_{\theta_1}(x_k) \) for \( k = 1, \dots, n \). Then, the \( k \)-th block of \( \mathbf{f}_{\theta_1} \) is
\begin{align}
    [\mathbf{f}_{\theta_1}]_k & = f_{\theta_1}(x_k) \\& = [\mathbf{f}_{\theta_0}]_k - \nonumber \eta \sum_{j=1}^n \Theta_{\theta_0}(x_k, x_j) [\mathbf{f}_{\theta_0}]_j \\ & \quad \ + \eta \sum_{j=1}^n \Theta_{\theta_0}(x_k, x_j) y_j.
\end{align}
In matrix form, we get that the term \( \sum_{j=1}^n \Theta_{\theta_0}(x_k, x_j) [\mathbf{f}_{\theta_0}]_j \) is the \( k \)th block of the matrix-vector product \( \mathbf{\Theta} \mathbf{f}_{\theta_0} \). Similarly, \( \sum_{j=1}^n \Theta_{\theta_0}(x_k, x_j) y_j \) is the \( k \)th block of \( \mathbf{\Theta} \mathbf{y} \).
Thus, the full vector updating $\mathbf{f}_{\theta_1} - \mathbf{f}_{\theta_0}$ satisfies
\begin{align}
\mathbf{f}_{\theta_1} = \mathbf{f}_{\theta_0} - \eta \mathbf{\Theta} \mathbf{f}_{\theta_0} + \eta \mathbf{\Theta} \mathbf{y} = (I - \eta \mathbf{\Theta}) \mathbf{f}_{\theta_0} + \eta \mathbf{\Theta} \mathbf{y}.
\end{align}
Let \( P = I - \eta \mathbf{\Theta} \) and \( \mathbf{b} = \eta \mathbf{\Theta} \mathbf{y} \), we obtain
\[
\mathbf{f}_{\theta_1} = P \mathbf{f}_{\theta_0} + \mathbf{b}.
\]
This is an affine transformation in \( \mathbb{R}^{n  d} \), which is parameterized by the real-valued matrix \( P \) and vector \( \mathbf{b} \).
\end{proof}

\begin{figure*}[htbp]
    \centering
    \begin{subfigure}[b]{1.0\linewidth}
        \centering
        \includegraphics[width=\textwidth]{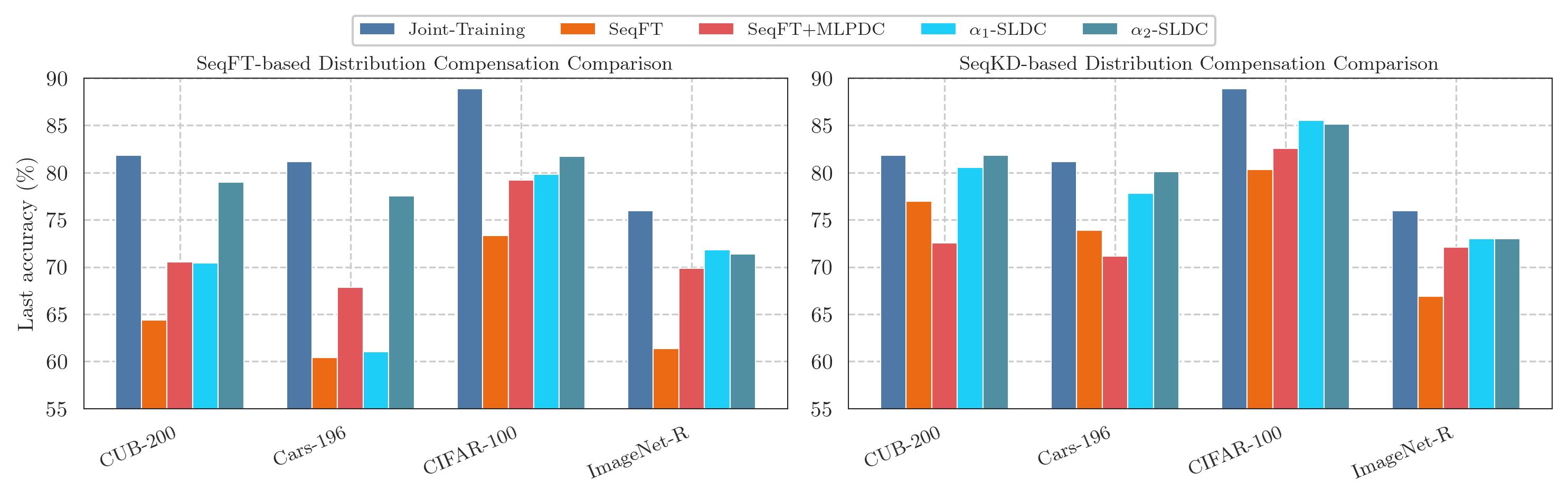}
        \caption{Performance comparison on MoCo-V3 pre-trained ViT-B/16 without ADE}
        \label{fig:moco_results_a}
    \end{subfigure}
    
    \begin{subfigure}[b]{1.0\linewidth}
        \centering
        \includegraphics[width=\textwidth]{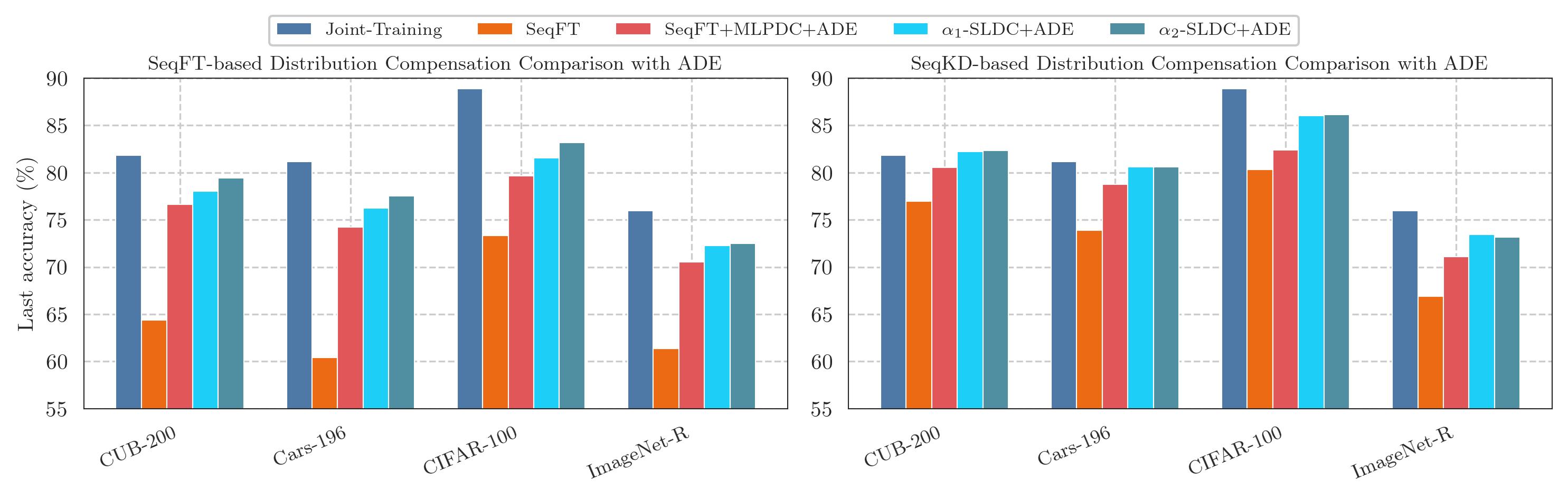}
        \caption{Performance comparison on MoCo-V3 pre-trained ViT-B/16 with ADE}
        \label{fig:moco_results_b}
    \end{subfigure}
    \caption{Comparative evaluation of CIL methods using self-supervised (MoCo-V3) pre-trained ViT-B/16 backbone. (a) Results without auxiliary data enrichment (ADE); (b) Results with ADE. (a) Performance comparison of SeqFT-based distribution compensation methods. (b) Performance comparison of SeqKD-based distribution compensation methods. Particularly, the results of joint-training serve as the performance upper bound for other mehtods.}
    \label{fig:moco_results}
\end{figure*}

\begin{figure*}[htbp]
    \centering
    \begin{subfigure}[b]{1.0\linewidth}
        \centering
        \includegraphics[width=\textwidth]{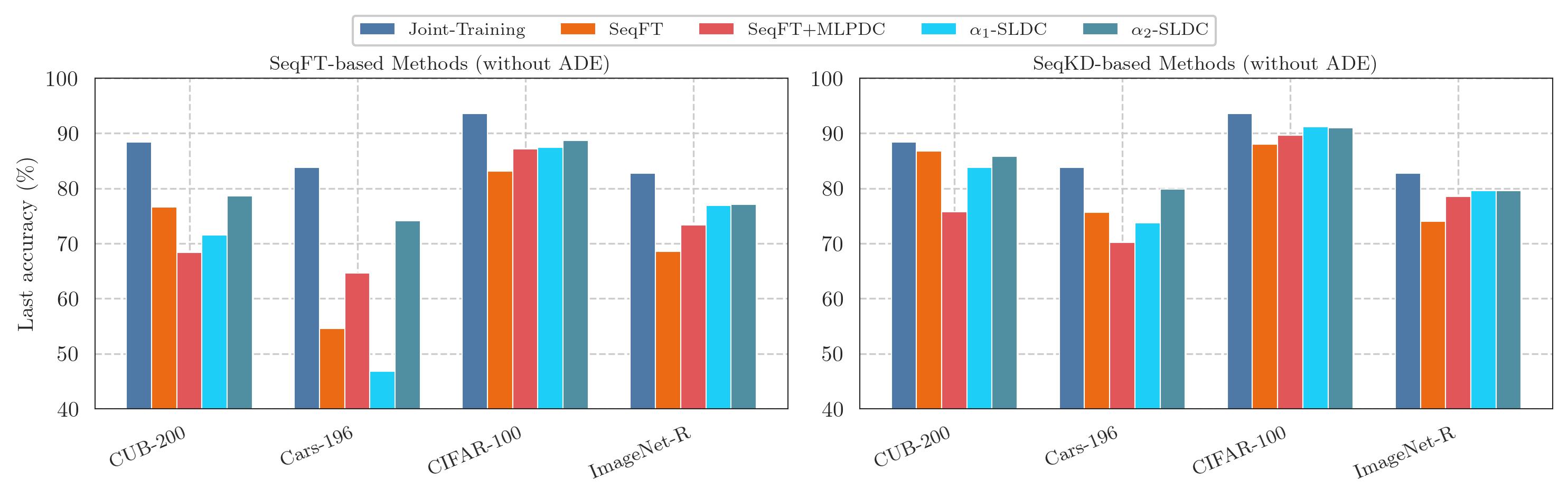}
        \caption{Performance comparison on ImageNet-21K pre-trained ViT-B/16 without ADE}
        \label{fig:sup21k_results_a}
    \end{subfigure}
    
    \begin{subfigure}[b]{1.0\linewidth}
        \centering
        \includegraphics[width=\textwidth]{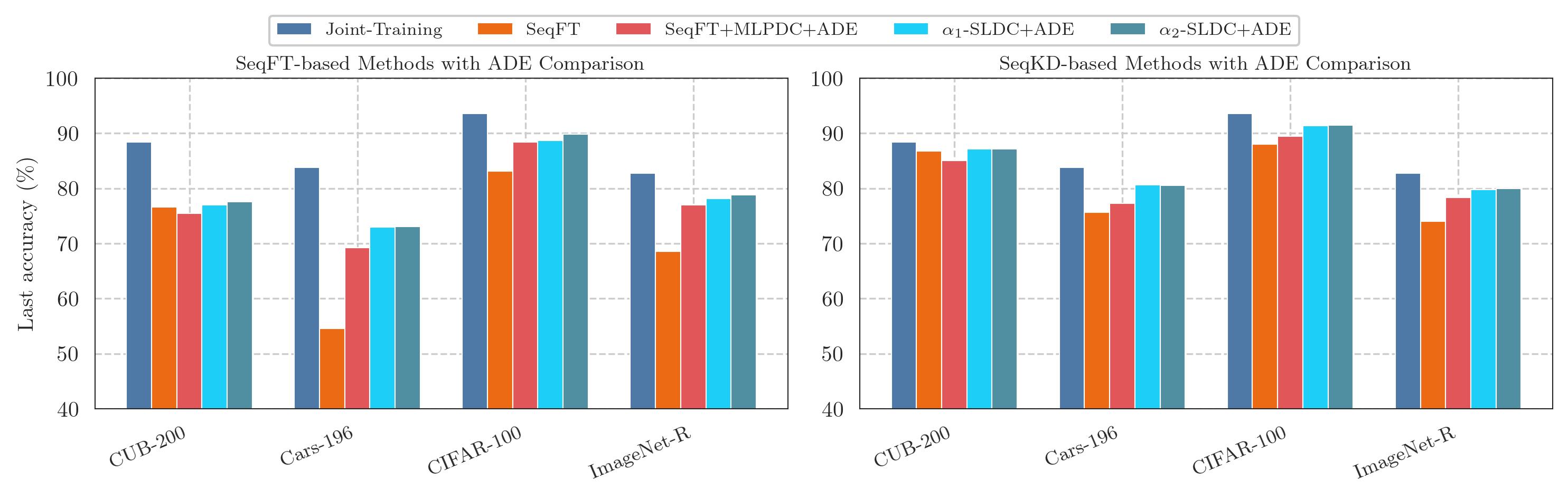}
        \caption{Performance comparison on ImageNet-21K pre-trained ViT-B/16 with ADE}
        \label{fig:sup21k_results_b}
    \end{subfigure}
    \caption{Comparative evaluation of CIL methods using supervised (ImageNet-21K) pre-trained ViT-B/16 backbone. (a) Results without auxiliary data enrichment (ADE); (b) Results with ADE. (a) Performance comparison of SeqFT-based distribution compensation methods. (b) Performance comparison of SeqKD-based distribution compensation methods. Particularly, the results of joint-training serve as the performance upper bound for other mehtods.}
    \label{fig:sup21k_results}
\end{figure*}

\begin{figure*}[htbp]
    \centering
    \includegraphics[width=0.9\linewidth]{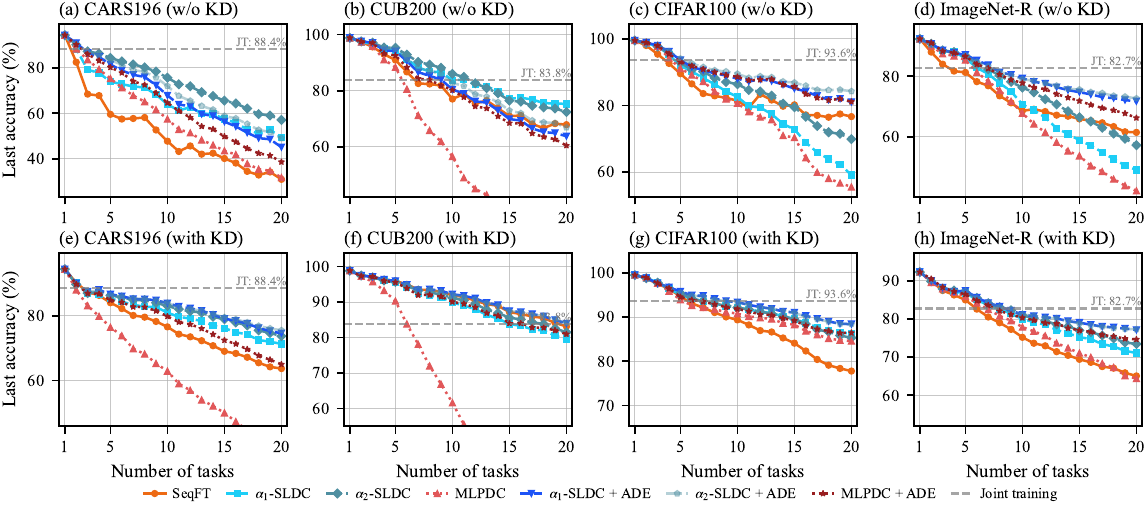}
    \caption{Performance comparison of SLDC methods on a 20-task sequence, demonstrating state-of-the-art results both with and without knowledge distillation.}
    \label{fig:sldc-long-sequence}
\end{figure*}

\end{document}